%% file: paper.tex
\documentclass[]{fairmeta}
\usepackage{natbib}
\usepackage{amssymb}
\usepackage{microtype}
\usepackage{hyperref}
\usepackage{url}
\usepackage{booktabs}
\usepackage{amsmath}
\usepackage{braket}
\usepackage{lineno}
\usepackage{graphicx}
\usepackage{subcaption}
\usepackage{wrapfig}
\usepackage{amsthm}
\usepackage{tcolorbox}
\usepackage{enumitem}
\newtheorem{proposition}{Proposition}
\newtheorem{theorem}{Theorem}
\usepackage{booktabs}

\input{macros}

\definecolor{darkblue}{rgb}{0, 0, 0.5}
\hypersetup{colorlinks=true, citecolor=darkblue, linkcolor=darkblue, urlcolor=darkblue}

\title{Distilled Pretraining: A modern lens of Data, \\ In-Context Learning and Test-Time Scaling}

\author[1,2,*]{Sachin Goyal}
\author[1]{David Lopez-Paz}
\author[1]{Kartik Ahuja}
\affiliation[1]{FAIR at Meta}
\affiliation[2]{Carnegie Mellon University}

\contribution[*]{Work done at Meta}

\abstract{In the past year, distillation has seen a renewed prominence in large language model (LLM) \emph{pretraining}, exemplified by the Llama-3.2 and Gemma model families. While distillation has historically been shown to improve statistical modeling, its effects on new paradigms key to modern LLMs—such as \emph{test-time scaling} and \emph{in-context learning}—remain underexplored. 
In this work, we make three main contributions.
First, we show that pretraining with distillation yields models that exhibit remarkably better test-time scaling.
Second, we observe that this benefit comes with a trade-off: distillation impairs in-context learning capabilities, particularly the one modeled via induction heads. 
Third, to demystify these findings, we study distilled pretraining in a sandbox of a bigram model, which helps us isolate the common principal factor behind our observations.
Finally, using these insights, we shed light on various design choices for pretraining that should help practitioners going forward.
}

\date{\today}
\correspondence{Sachin Goyal at sach.goyalsachin@gmail.com}

\begin{document}

\maketitle

\section{Introduction}

Knowledge distillation, first proposed by \cite{buciluǎ2006model} for compressing ensembles, was later popularized by seminal works of \cite{ba2014deep} and \cite{hinton2015distilling}.
However, distillation didn’t trickle into the pipelines of early large language models (LLMs)—such as GPT-2/3 and Llama 1/2. But more recently, distillation has resurged as a prominent method in the LLM landscape, not just during post-training, but also \emph{pretraining} as seen in the Llama-3.2~\citep{meta2024llama32} and Gemma~\citep{team2024gemma,team2025gemma}. This shift reflects a growing reality: extremely large models (e.g., Llama-4-Behemoth~\citep{llama4meta}) are too costly to deploy widely and will increasingly serve solely as teachers for distilling smaller, more practical models. Going forward, these deployed models are likely to be pretrained entirely via distillation as seen in Llama-4-Maverick~\citep{llama4meta} that was distilled from Llama-4-Behemoth.

Despite its growing role, the science of distillation (using soft labels) in modern LLM \emph{pretraining} has remained largely unexplored.
Gemma-3 and Llama-3.2 models show clear empirical benefits on standard benchmarks from pretraining with distillation. However, these models typically leverage teachers trained on far more data than the students. This raises a fundamental question: are the gains from distillation merely a result of additional teacher data, or do they reflect unique benefits beyond extra data exposure?
As we hit the data wall, will distillation continue to be beneficial? Moreover, modern LLMs are no longer limited to evaluation on standard benchmarks. \emph{New paradigms such as in-context learning and test-time scaling are key to current LLM frontiers, yet the effect of pretraining with distillation on these paradigms remains largely unexamined}.

\begin{figure}[t!]
    \centering
    \begin{subfigure}{0.315\textwidth}
        \centering
        \includegraphics[width=\linewidth]{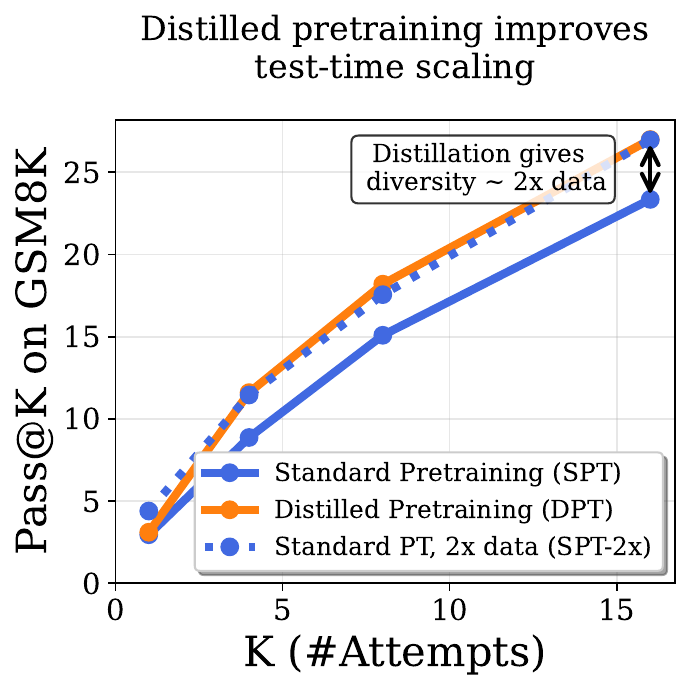}
        \caption{}
    \end{subfigure}
    \hfill
    \begin{subfigure}{0.325\textwidth}
        \centering
        \includegraphics[width=\linewidth]{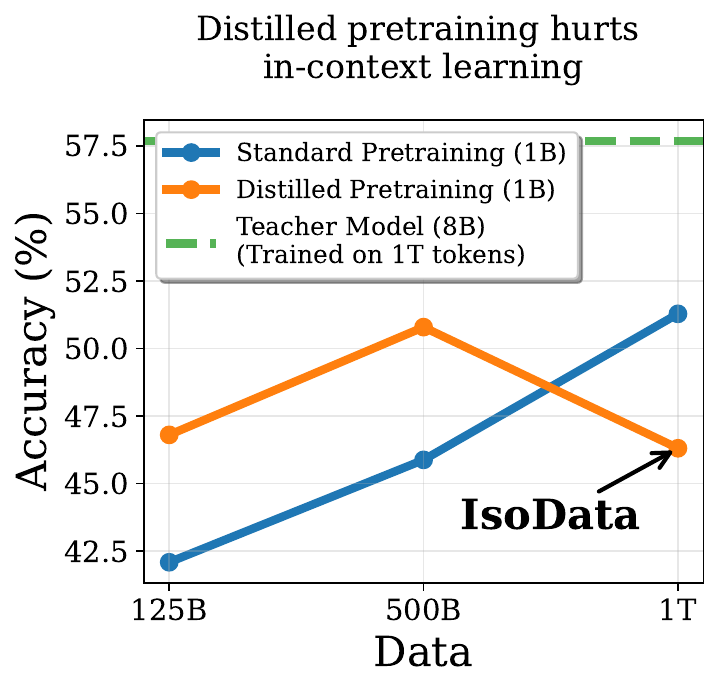}
        \caption{}
    \end{subfigure}
    \hfill
    \begin{subfigure}{0.315\textwidth}
        \centering
        \includegraphics[width=\linewidth]{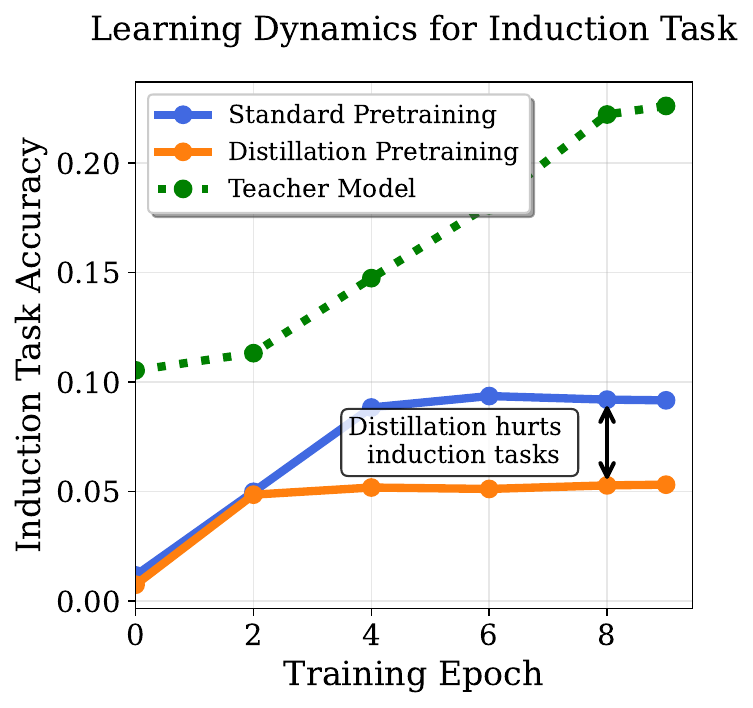}
        \caption{}
    \end{subfigure}
    \caption{\textbf{Distilled pretraining in modern LLM regime} \textbf{(a)} Comparing standard pretraining (SPT) with distilled pretraining (DPT). On reasoning tasks like GSM8k, although both the models have a similar \passone{}, DPT substantially outperforms SPT on \passk{} for higher $k$ (27\% vs 23\% for $k=16$). Infact, DPT matches the \passsixteen{} of a standard pretrained model trained on twice the data. \textbf{(b)} Distilled pretraining hurts in-context learning capabilities when the student and teacher model see the same data. In the figure, as we scale the student data to 1T (data seen by the teacher), the gains of distillation over standard pretraining on in-context learning tasks diminish (Figure~\ref{fig:isodata_icl} for details). \textbf{(c)} We demystify these findings by analyzing a bigram sandbox, where we show that training with distillation can impair the learning of induction heads~\citep{bietti2023birth}, which form the key mechanism behind in-context learning.
    }
    \label{fig:title_base_model_passk}
\end{figure}

In this work, we uncover key trade-offs associated with distilled pretraining (DPT). First, we show that DPT remains beneficial on standard language modeling tasks, even in the data-constrained regime where the student and the teacher models are trained on the same data. This suggests promise for scaling DPT further. However, in contrast, we observe that \emph{naively scaling pretraining with distillation (DPT) hurts the in-context learning performance} (Figure~\ref{fig:title_base_model_passk}b). In particular, distillation impairs the learning of induction heads~\citep{olsson2022incontextlearninginductionheads}—the transformer circuits that enable models to search and copy from context(Figure~\ref{fig:title_base_model_passk}c).

Strikingly, the very process of distillation that undermines in-context learning, at the same time also yields models that demonstrate \emph{markedly better test-time scaling capabilities}. We study this through \passk{}, where the model is allowed multiple attempts per question. Distilled models outperform standard pretraining on \passk{} at larger $k$, even when \passone{} is the same (Fig.\ref{fig:title_base_model_passk}a). On GSM8k, for example, both models have the same \passone, but the distilled model achieves a much higher \passsixteen{}—27\% versus 23\%. Remarkably, it even matches the \passsixteen{} of a standard-pretrained model trained on twice the data, despite a lower \passone. Similar patterns hold on MATH and MBPP, where distilled pretraining consistently improves test-time scaling by enhancing generation diversity\citep{dang2025weightensemblingimprovesreasoning}.

Interestingly, the mechanisms through which distillation undermines in-context learning are the same ones that enhance test-time scaling. 
We study this tradeoff in a simple yet expressive sandbox of a bigram model~\citep{bietti2023birth, edelman2024evolutionstatisticalinductionheads}. 
A bigram model is characterized by a matrix in which each row represents the next token probability distribution over the vocabulary.
Pretraining with distillation is beneficial in learning the high-entropy rows. These rows basically model prompts like ``I work at'', which admit multiple valid completions (e.g., ``gym'', ``hospital'', ``restaurant'').  In contrast, distillation does not help in learning low-entropy rows which model the deterministic state transitions (prompts), e.g., induction heads where the next-token probability distribution is one-hot. For these cases, distillation does not provide any information beyond what is already there in ground truth one-hot labels. Worse, an imperfect teacher can hurt the learning of these low-entropy rows by introducing noise via soft probability distribution(Figure~\ref{fig:title_base_model_passk}c). 

Finally, borrowing insights from our analysis, we discuss various design choices for improving pretraining. These include \emph{distillation-specific data curation}, teacher selection, and comparisons with other recent advances such as multi-token prediction~\citep{gloeckle2024betterfasterlarge}, which we hope will aid practitioners going forward. We summarize our key contributions in this work below:
\begin{itemize}
    \item \textbf{Test-time scaling:} We show that distilled pretraining produces models with markedly stronger test-time scaling, often matching standard pretraining on up to twice the data.  
    \item \textbf{In-context learning trade-off:} We find that these gains come at a cost, as distillation impairs in-context learning, particularly by weakening induction heads.  
    \item \textbf{Bigram analysis:} We isolate the common mechanism that drives the improvements in test-time scaling but impairs in-context learning at the same time.
    \item \textbf{Practitioner Takeaways:} We translate these insights into concrete design choices for improving pretraining with distillation, including distillation specific data curation, teacher selection, etc.
\end{itemize}

\subsection{Preliminaries}

We start by revisiting the setup of distillation from \cite{hinton2015distilling}.  We are given a dataset $\{(x_i,y_i)\}_{i=1}^{n}$ of inputs $x_i \in \mathbb{R}^{d}$'s and the labels $y_i's \in \Delta^{k-1}$, where $k$ is the number of classes and $\Delta^{k-1}$ is a probability simplex over those classes.  Let us begin with the objective of training a model from scratch on the above data using cross-entropy loss $\ell$, $h^{\star} \ \in \arg\min_{h \in \mathcal{H}} \frac{1}{n}\sum_{i=1}^{n}\ell\big(y_i, \sigma(h(x_i))\big),$ where $h$ is a candidate function drawn from the hypothesis class $\mathcal{H}$, 
$\sigma:\mathbb{R}^{k}\rightarrow \Delta^{k-1}$ is the softmax function $\sigma_j(z) = \frac{\exp(z_j)}{\sum_{i=1}^{k} \exp(z_i)}$ and $\ell(y,\hat{y}) = -\sum_{j=1}^{k} y_j \log(\hat{y}_j)$. 

We are now ready to define the standard objective used in distillation:
\begin{equation}
    h^{\dagger} \in \arg\min_{h \in \mathcal{H}} \frac{1}{n}\Big[(1-\alpha)\sum_{i=1}^{n}\ell\big(y_i, \sigma(h(x_i))\big) + \alpha \sum_{i=1}^{n}\ell\big(s_i, \sigma(h(x_i))\big)\Big],
    \label{eqn:erm_distillation}
\end{equation}

where $\alpha \in [0,1]$ and  $s_i= \sigma(h_{\mathsf{teacher}}(x_i)/T)$ is a soft label generated by the teacher using a temperature $T$.

This form of distillation has recently been adopted in pretraining language models as well. We first start by describing the next-token prediction objective over a sequence $(x_1, \cdots, x_t)$

\begin{equation}
   \frac{1}{t}\sum_{j=1}^{t} \ell\Big(x_{j+1}, \sigma(h(x_{\leq j}))\Big)
\end{equation}

The objective function used in pretraining distillation is

\begin{equation}
   \frac{1}{t}\Big[\sum_{j=1}^{t-1} (1-\alpha)\ell\Big(x_{j+1}, \sigma(h(x_{\leq j}))\Big) + \alpha \sum_{j=1}^{t-1}\ell\Big(s_{j+1}, \sigma(h(x_{\leq j}))\Big)\Big], 
\end{equation}
where  $s_{j+1}= \sigma(h_{\mathsf{teacher}}(x_{\leq j})/T)$

\section{No Extra Data: Does Distillation Still Improve Performance?}
\label{sec:isodata}

\begin{figure}[t!]
    \centering
    \begin{subfigure}{0.24\textwidth}
        \centering
        \includegraphics[width=\linewidth]{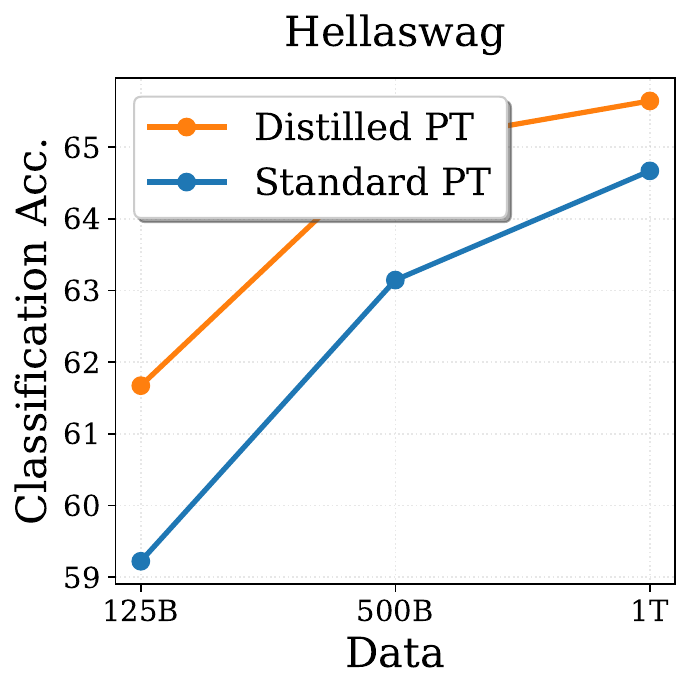}
    \end{subfigure}
    \hfill
    \begin{subfigure}{0.24\textwidth}
        \centering
        \includegraphics[width=\linewidth]{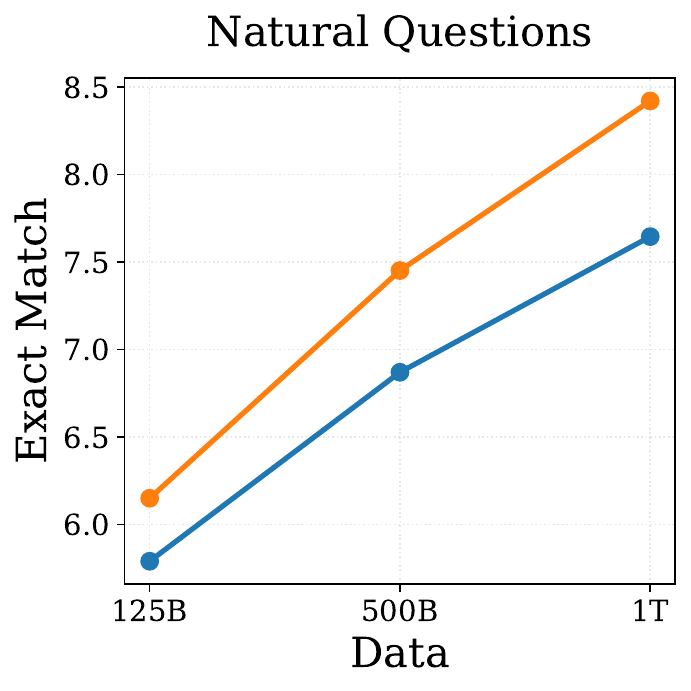}
    \end{subfigure}
    \hfill
    \hfill
    \begin{subfigure}{0.245\textwidth}
        \centering
        \includegraphics[width=\linewidth]{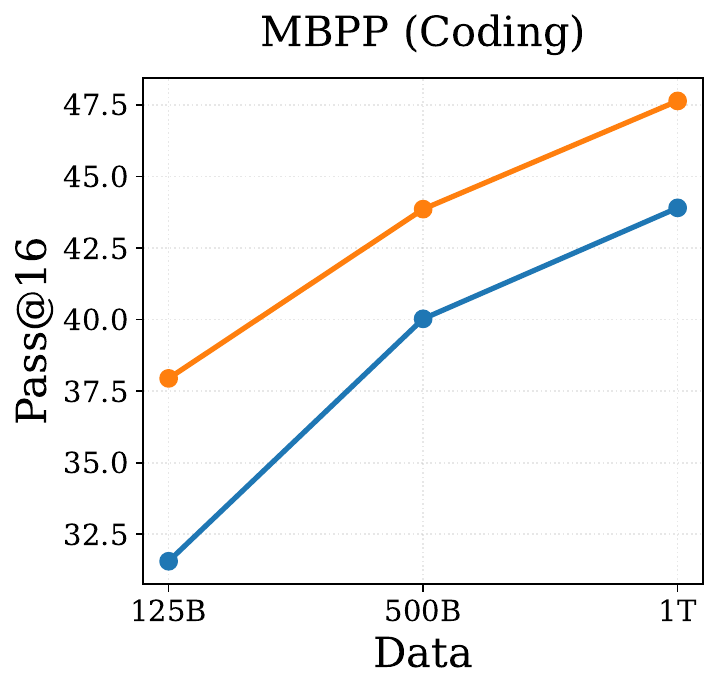}
    \end{subfigure}
    \hfill
    \begin{subfigure}{0.235\textwidth}
        \centering
        \includegraphics[width=\linewidth]{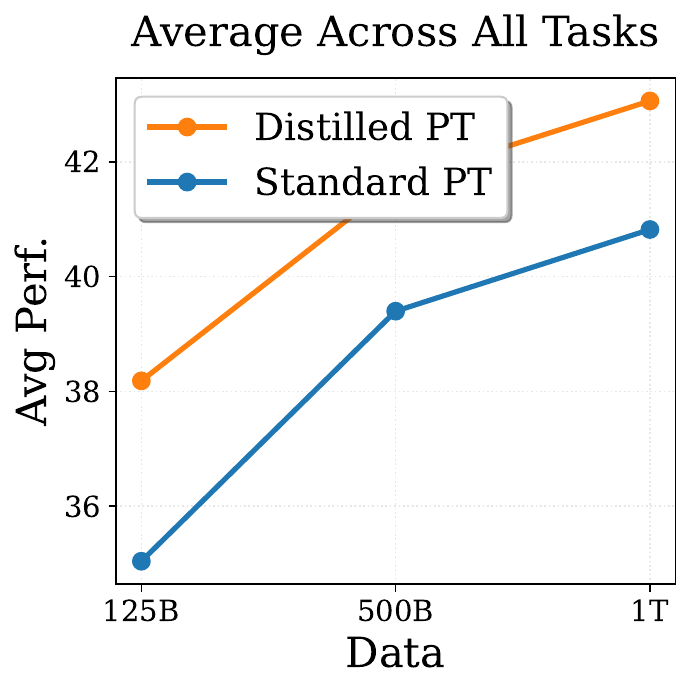}
    \end{subfigure}
    \caption{\textbf{IsoData Distillation (\S~\ref{sec:isodata})}: Will distilled pretraining remain effective when the student and teacher are trained on the same data? To explore this, we use an 8B model trained on 1T tokens as a teacher. Using this teacher, we train various student models, with and without distillation, scaling up the data to the exact same 1T tokens. We observe that even in the IsoData case where both teacher and student have seen the same 1T tokens, the distilled model generally  outperforms standard pretraining on standard language modeling tasks. Thus distillation generally remains beneficial even in a data-constrained regime. See Figure~\ref{fig:isodata_iwl_full} for more tasks.
    }
    \label{fig:isodata_iwl}
\end{figure}

Recent pretrained LLM families—such as the Gemma-3 and Llama-3.2 series—have shown clear benefits from distillation compared to training from scratch. However, these models typically leverage teachers trained on significantly more data than the students ultimately use, raising a fundamental yet unanswered question: Are the gains from distillation simply due to this additional teacher data? Or does distillation offer unique benefits beyond merely seeing extra data via the teacher?

We begin this work, by first answering the basic question raised above via a set of ``IsoData Distillation'' experiments. 
We first train an 8B teacher model on 1T tokens.
We then train 1B students—with and without distillation—on the same 1T tokens to see if distillation still helps when both see identical data.
Figures~\ref{fig:isodata_iwl} and ~\ref{fig:isodata_iwl_full} compare the performance of the two 1B models on the standard language modeling tasks like COPA, HellaSwag, NaturalQA, TQA, GSM8k, etc. We observe that distillation continues to generally benefit even when the training is scaled to the same data as seen by the teacher (1T tokens).

As a side note, we clarify that recent works like ~\citet{gu2025miniplmknowledgedistillationpretraining, busbridge2025distillationscalinglaws} have argued that under compute-matched conditions distillation offers no advantage (e.g., by giving standard pretraining more data or training steps to offset the computational cost of distillation).
However, these claims do not reflect many practical setups. First, teacher logits can be obtained by distributing inference on
more widely available, lower-cost GPUs (e.g., single GPUs without interconnect). 
Second, nowadays the teacher logits are cached during the training run of the teacher model itself. The logits from initial phase of the training are discarded though.
We believe it is more important to question the advantage of distillation under the data constrained scenarios as going forward, the compute-per-token budget will keep increasing as we hit the data wall.

\begin{tcolorbox}[left=0mm,right=0mm,colback=metablue!8!white,colframe=white]
Distilled pretraining (DPT) continues to be generally beneficial even in the data-constrained regime, when the student is shown the same amount of data as the teacher.
\end{tcolorbox}

In the next section, we will analyze distilled pretraining on new paradigms  centric to modern LLMs, beyond the standard language modeling tasks: in-context learning and test-time scaling.

\paragraph{Theoretical works on IsoData distillation:}
Theoretical analyses of distillation have primarily explained its benefits through two lenses: sample complexity and optimization. From the sample complexity perspective, \citet{menon2021statistical} show that distillation improves generalization when the teacher has access to more data (e.g., a Bayes-optimal teacher). However, this framework falls short in the IsoData regime, where teacher and student train on the same data. From the optimization perspective, \citet{safaryan2023knowledge} argue that distillation enables the student to converge closer to the Bayes-optimal solution as the teacher improves. Yet, it remains unclear whether such convergence is faster than that of standard SGD when no additional teacher data are available.

The only works in theory that explicitly address the IsoData setting have appeared only recently, and somewhat surprisingly. \citet{mobahi2020selfdistillationamplifiesregularizationhilbert} show that self-distillation can reduce overfitting by dampening variance along the top singular directions of the learned representation. Building on this, \citet{nagarajan2024studentteacherdeviationsdistillationdoes} demonstrate that distillation further exaggerates the implicit bias of gradient descent, driving the student to converge more rapidly along top eigendirections. Together, these results suggest that the gains from IsoData distillation arise less from sample complexity or optimization speedups, and more from implicit regularization effects acting through the singular spectrum of the representation.

\section{Distilled Pretraining Through the Modern Lens: In-Context Learning and Test-Time Scaling} 

Knowledge distillation has long been shown to improve \textit{in-weights learning} (IWL), resulting in stronger performance in standard evaluation tasks and benchmarks~\citep{team2024gemma,team2025gemma}. However, in modern LLMs, the desired capabilities extend much beyond the classical setting of IWL. 
The ability to generate \textit{diverse solution paths} is critical for skills like test-time scaling and search at inference~\citep{chow2024inferenceawarefinetuningbestofnsampling,dang2025weightensemblingimprovesreasoning,chen2025rethinkingfinetuningscalingtesttime}, but more crucially, to also enable better post-training with reinforcement learning with verifiable rewards (RLVR).
Likewise, \textit{in-context learning} (ICL)---where models learn and adapt from inference time prompts is especially desirable.

In this section, we examine how pretraining with distillation shapes these two capabilities key to the current frontier of LLMs: diversity for test-time scaling (as measured by \passk) and in-context learning (ICL). Through a series of controlled experiments, we isolate the effects of distillation and provide a comparative analysis across these dimensions.

\subsection{Distillation impairs in-context learning}
\label{sec:icl}

\begin{figure}[t!]
    \centering
    \begin{subfigure}{0.24\textwidth}
        \centering
        \includegraphics[width=\linewidth]{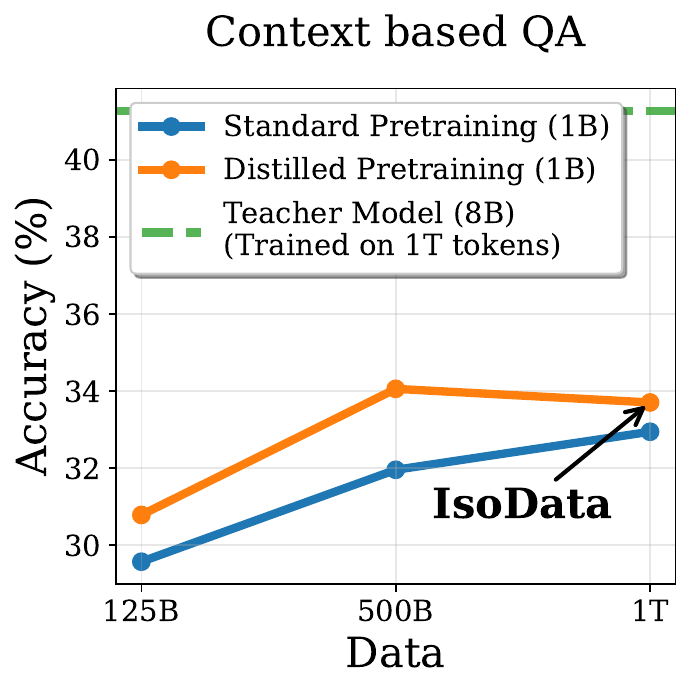}
        \caption{}
    \end{subfigure}
    \hfill
    \begin{subfigure}{0.24\textwidth}
        \centering
        \includegraphics[width=\linewidth]{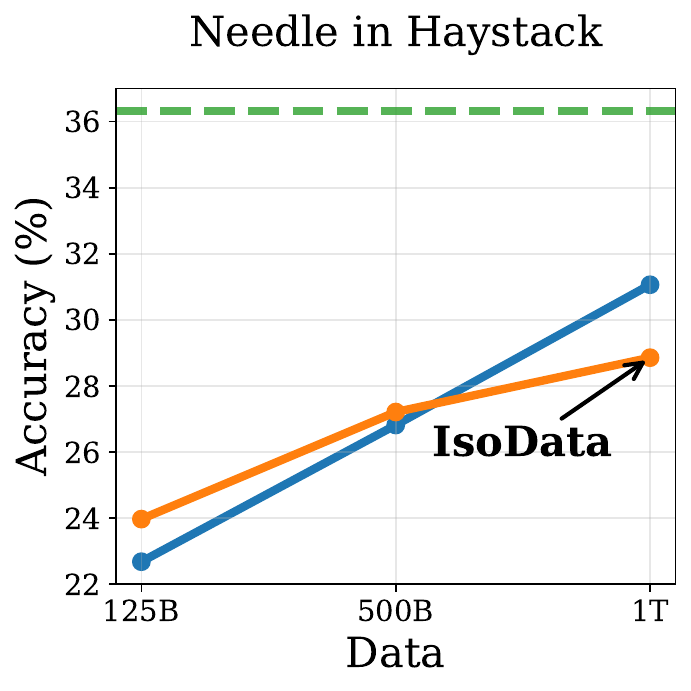}
        \caption{}
    \end{subfigure}
    \hfill
    \hfill
    \begin{subfigure}{0.24\textwidth}
        \centering
        \includegraphics[width=\linewidth]{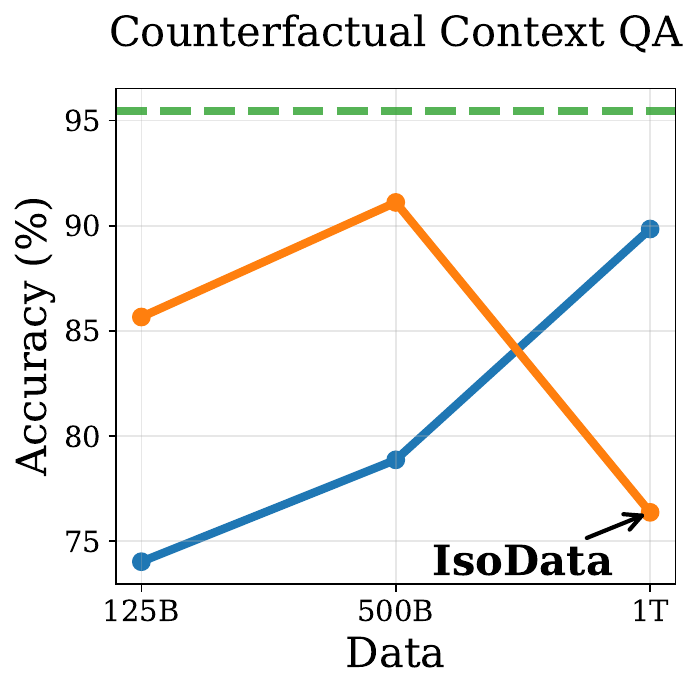}
        \caption{}
    \end{subfigure}
    \hfill
    \begin{subfigure}{0.24\textwidth}
        \centering
        \includegraphics[width=\linewidth]{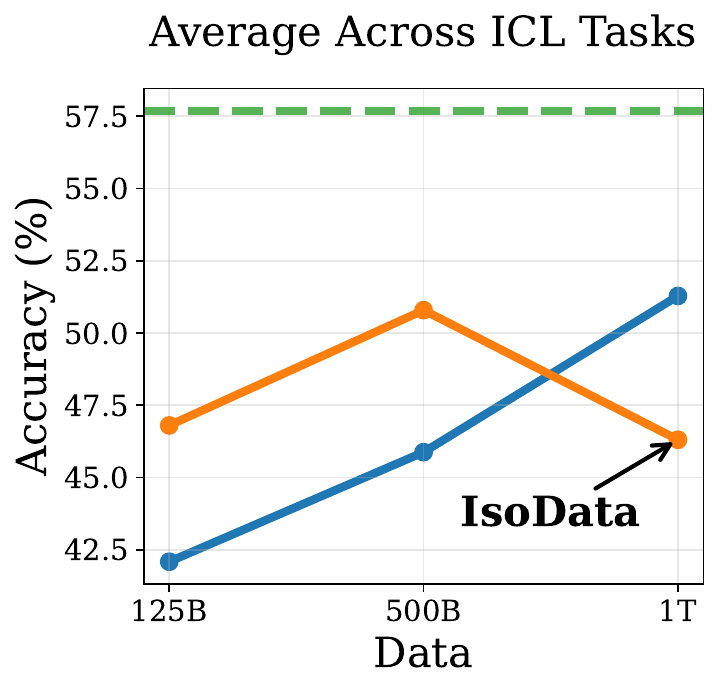}
        \caption{}
    \end{subfigure}
    \caption{\textbf{Distilled pretraining impairs in-context learning, especially in the IsoData setting (\S~\ref{sec:icl}): }We train 1B models with and without distillation using a 8B teacher trained on 1T tokens. We observe that the advantages of distillation on in-context learning tasks diminish as the amount of training data increases (each scatter is a separate model trained with a full LR scheduler). 
    Eventually, the distilled model underperforms in the IsoData setup, where both the teacher and student are trained on the same data.
    This is because induction heads which form a key mechanism behind in-context learning~\citep{olsson2022incontextlearninginductionheads} are built on low-entropy mappings, requiring the model to copy a specific token from earlier in the sequence. For these cases, distillation can't help—it can only match the hard label at best, and at worst, it actively hinders learning for such copying tasks by softening the supervision. 
    This is in contrast to performance on standard language modeling tasks where distillation continues to help even in IsoData setting (Figure~\ref{fig:isodata_iwl}). 
    }
    \label{fig:isodata_icl}
\end{figure}

The seminal work of ~\citet{olsson2022incontextlearninginductionheads} introduced induction heads as the key mechanism behind in-context learning in modern LLMs. Induction heads allow models to ``copy'' tokens from earlier positions in the input into later parts of the output~\citep{olsson2022incontextlearninginductionheads,edelman2024evolutionstatisticalinductionheads, bietti2023birth}. For example, if a prompt contains a token sequence such as ``I work at Gym,'' an induction head can help the model replicate ``Gym'' in a question about workplace that follows. Such copying ability is critical for tasks that require models to attend to and reuse information presented in the context.

\paragraph{Experimental Setup: }
We train an 8B teacher on 1T tokens, and then train 1B models—with and without distillation—scaling the data to the same set of 1T tokens as seen by the teacher (models are trained to convergence at each data scale). We call this as the ``IsoData'' setup where the student and the teacher models see the same data. This IsoData setup  is necessary to ensure a fair comparison, eliminating any indirect data advantage the distilled model might otherwise have. 
To evaluate model’s ability to copy from the context—a hallmark of induction heads that form the key mechanism behind in-context learning— we use 3 benchmarks: (a) context-based QA (DROP~\citep{Dua2019DROP}, RACE~\citep{lai-etal-2017-race}), where questions must be answered using the accompanying context; (b) needle-in-a-haystack task (babilong~\citep{kuratov2024babilong}), which requires locating sparse information embedded in long contexts; and (c) counterfactual context QA ~\citep{goyal2025contextparametricinversioninstructionfinetuning}, where the correct answer as per the context contradicts factual knowledge (i.e. answer based on model's memory or weights), forcing the model to rely solely on contextual cues. 

\paragraph{Observations: }Figure~\ref{fig:isodata_icl} compares the in-context learning performance of the two 1B models trained with and without distillation, 
as the training data is scaled to 1T tokens. 
We observe a consistent pattern that as the training tokens are increased, the relative advantage of distillation over the standard pretrained model keeps on diminishing. Infact the distilled model eventually underperforms in the IsoData setup (1T tokens) on needle-in-haystack and counterfactual-QA tasks. These observations are in stark contrast to the observations on standard language modeling tasks (e.g., Hellaswag, GSM8k, NaturalQA) in Figure~\ref{fig:isodata_iwl}, where distillation continues to offer advantage even in the ``IsoData'' setup. We note that the distilled model can be expected to perform better than standard pretrained model in the non-isodata setting (say 125B and 500B training in Figure~\ref{fig:isodata_icl}) because of the indirect data advantage through the teacher. For example, the teacher which has been trained on 1T tokens might have seen more variety of copying tasks and hence can give better supervision. Therefore, comparison under IsoData setup is crucial.

\paragraph{Why does distillation hurt in-context learning?}  
For inputs with deterministic outputs—such as “What is 2+3=”—a perfect teacher's soft labels reduce to the same one-hot, \emph{low-entropy} labels already present in ground truth. There’s no additional learning signal. Worse, real teachers are imperfect: their predictions often assign non-zero mass to distractors, subtly injecting noise into what should be a clean, unambiguous target.  

\begin{tcolorbox}[left=0mm,right=0mm,colback=metablue!8!white,colframe=white]
Induction tasks are built on low-entropy mappings, requiring the model to copy a specific token from earlier in the sequence. For these cases, distillation can't help—it can only match the hard label at best. Worse, it can actively hinder learning for such copying tasks by softening the supervision. 
\end{tcolorbox}

We discuss this phenomenon more formally in our bigram sandbox in \S~\ref{sec:bigram}. Finally, in \S~\ref{sec:token_routing}, we discuss some initial mitigation strategies motivated by our sandbox.

\subsection{Distillation helps diversity}
\label{sec:distillation_improves_diversity}

\begin{figure}[t!]
    \centering
    \begin{subfigure}{0.32\textwidth}
        \centering
        \includegraphics[width=\linewidth]{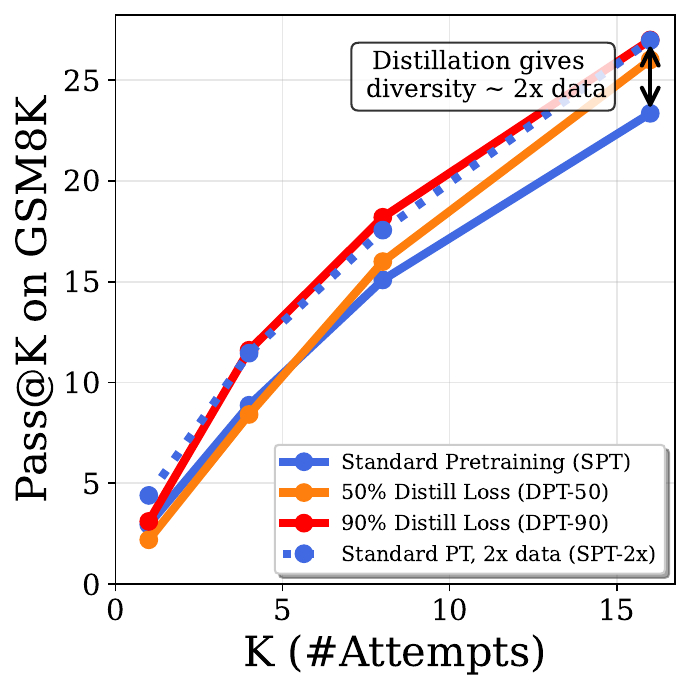}
        \caption{}
    \end{subfigure}
    \hfill
    \begin{subfigure}{0.32\textwidth}
        \centering
        \includegraphics[width=\linewidth]{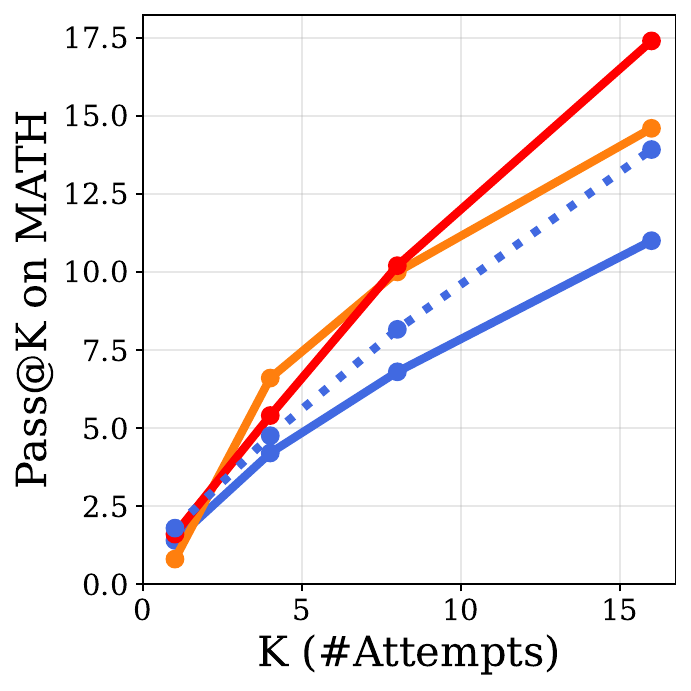}
        \caption{}
    \end{subfigure}
    \hfill
    \begin{subfigure}{0.32\textwidth}
        \centering
        \includegraphics[width=\linewidth]{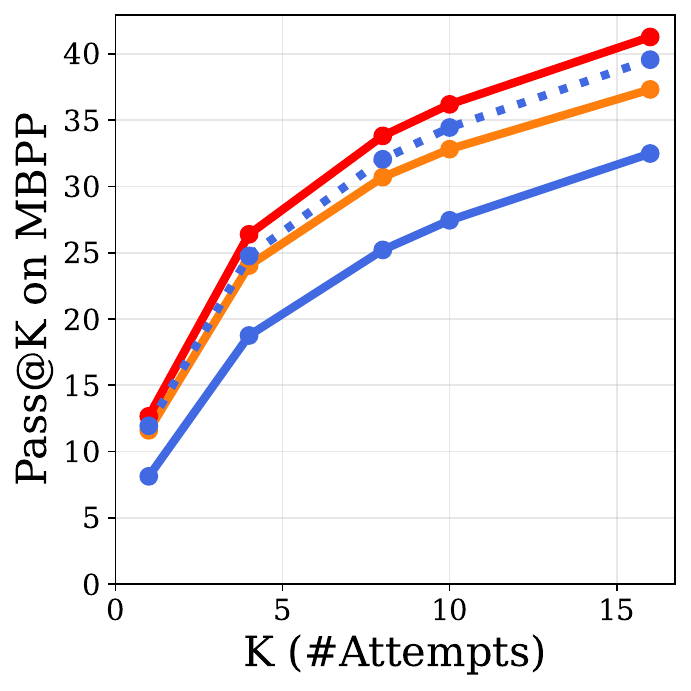}
        \caption{}
    \end{subfigure}
    \hfill
    \begin{subfigure}{0.32\textwidth}
        \centering
        \includegraphics[width=\linewidth]{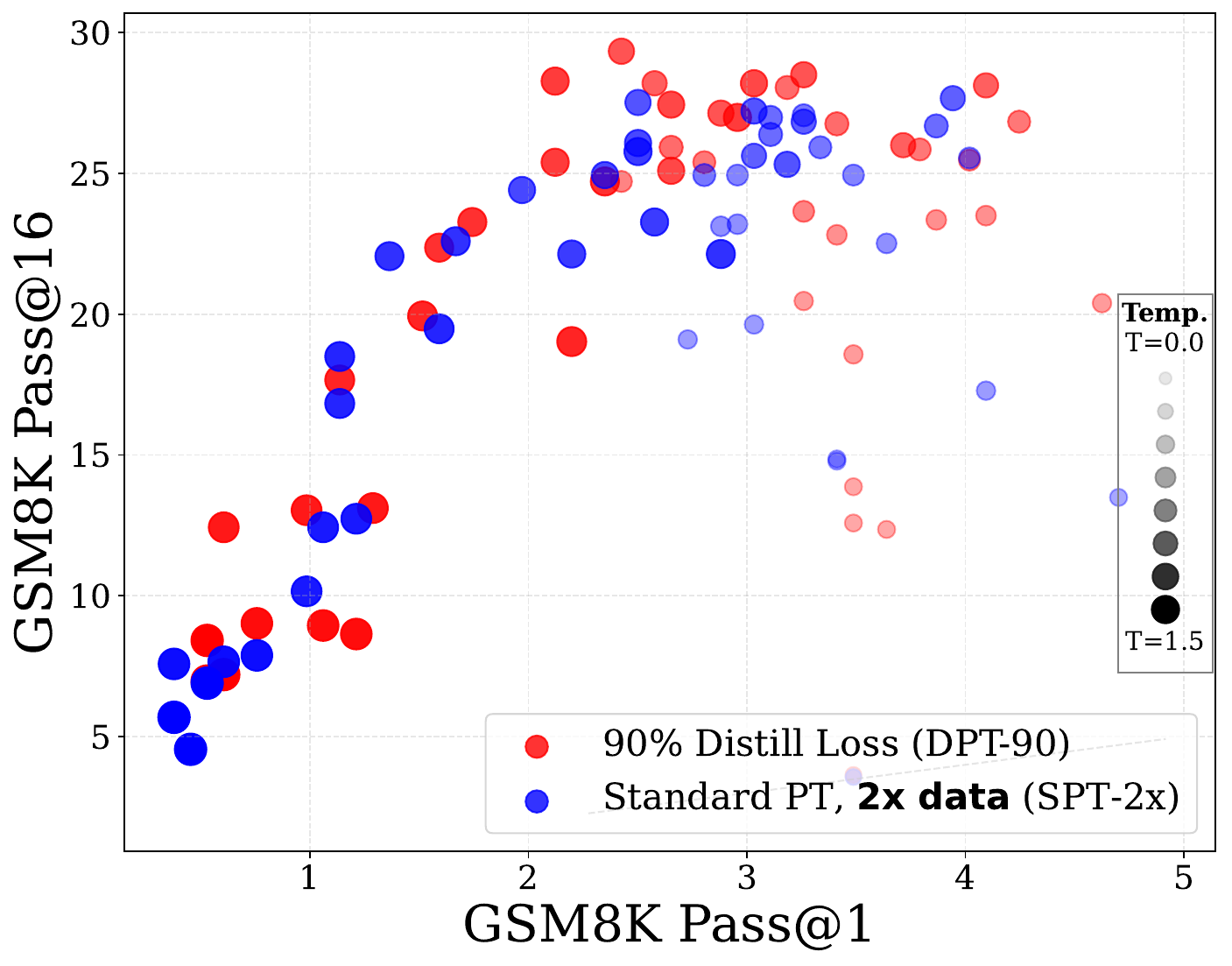}
        \caption{}
    \end{subfigure}
    \hfill
    \begin{subfigure}{0.32\textwidth}
        \centering
        \includegraphics[width=\linewidth]{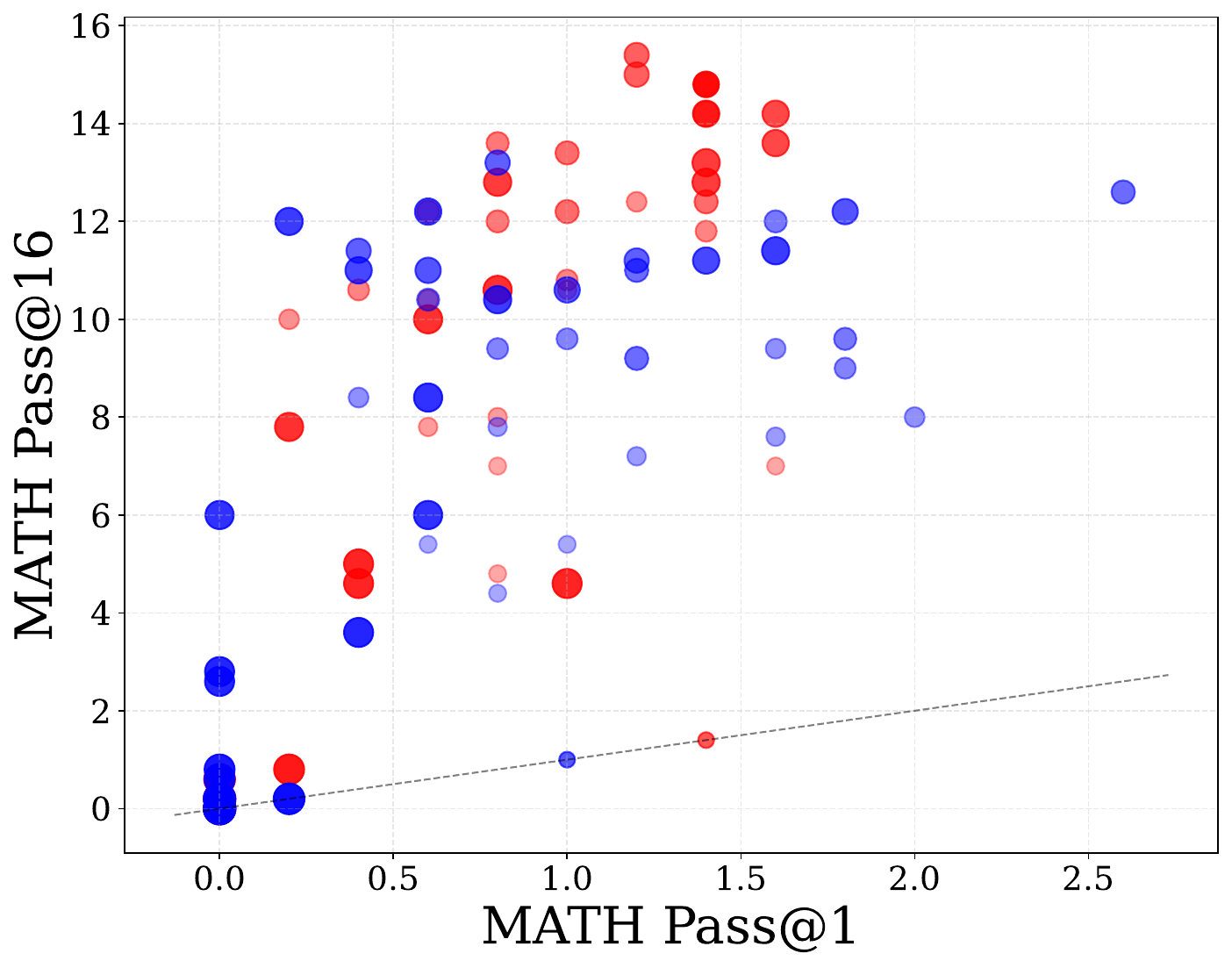}
        \caption{}
    \end{subfigure}
    \hfill
    \begin{subfigure}{0.32\textwidth}
        \centering
        \includegraphics[width=\linewidth]{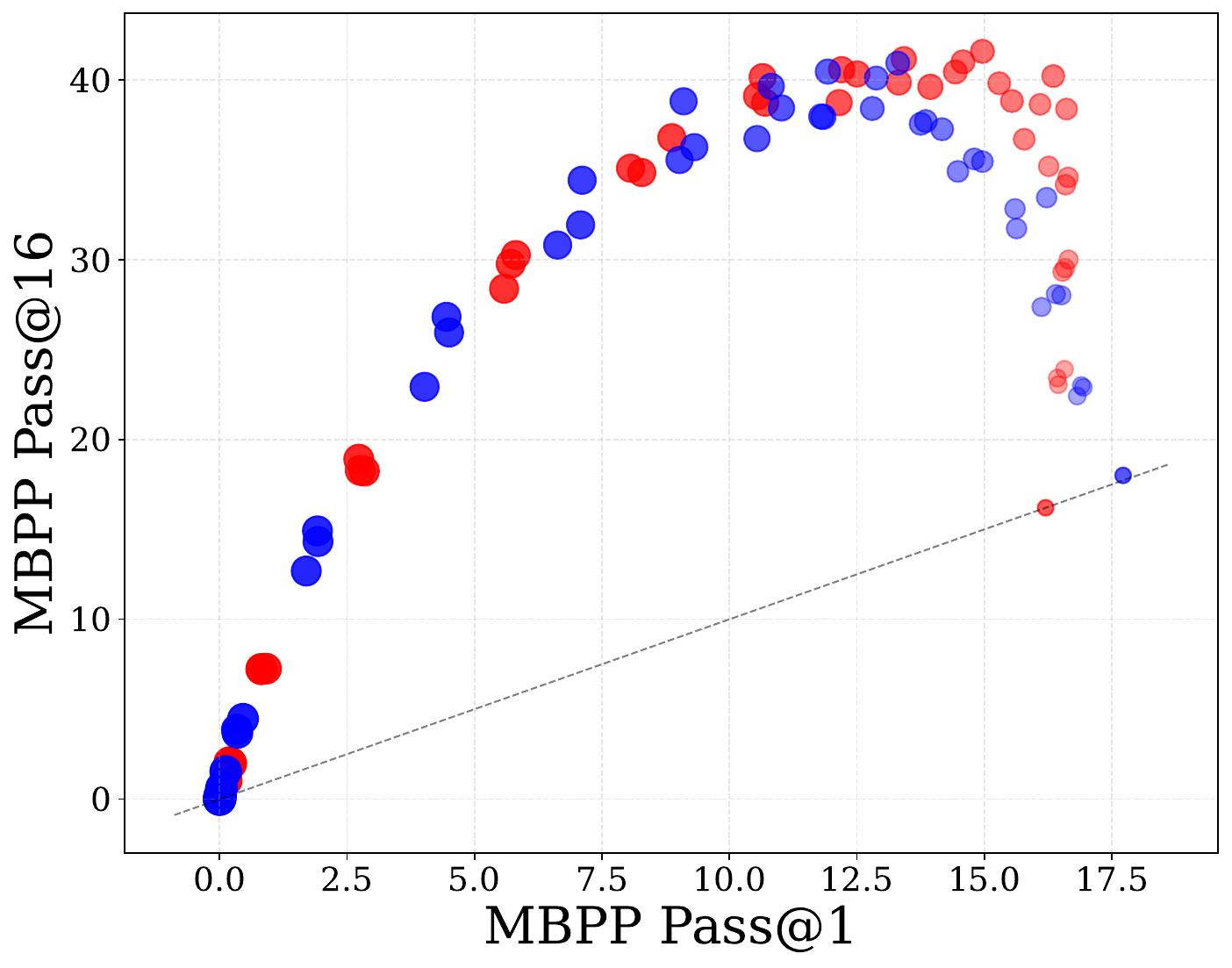}
        \caption{}
    \end{subfigure}
    \caption{\textbf{Distilled pretraining improves generation diversity and enables superior test-time scaling(\S~\ref{sec:distillation_improves_diversity}): }
    \textbf{Top-row (a-c): }We plot \passk{} curves with temperature first optimized for \passsixteen{} performance.
    Distilled pretraining with 50\% weight of distillation (DPT-50) consistently outperforms standard pretraining (SPT) on \passsixteen{}, even though it has worse \passone{} on GSM8k and MATH.
    \textbf{Bottom-row (d-f): }Next we increase the weight of distillation during pretraining to 90\% (DPT-90) and compare it to  a harder baseline of standard pretraining on \emph{2x data} (SPT-2x). We plot \passone{} vs \passsixteen{}, at temperatures from 0 to 1.5 with a granularity of 0.1. Observe that the DPT-90 model generally has a higher \passsixteen{} given any \passone{}, despite being trained on half the data. Infact, the DPT-90 model has the highest \passsixteen{} (lies at the top) on all three benchmarks, despite having a similar or lower \passone{} (lies towards left) compared to the SPT-2x model, when searching over various temperatures and seeds.
    This indicates that distilled pretraining rewards with models that showcase higher diversity in generations and better test-time scaling trends.  }
    \label{fig:test_time_scaling_main}
\end{figure}

\paragraph{Experimental Setup:}
We train 1B models on 125B tokens, with and without distillation, using the Llama-3.1-8B base model as the teacher. For distilled pretraining (DPT), we consider two settings: DPT-50 and DPT-90, where the distillation loss is weighted at 50\% and 90\% respectively (see $\alpha$ in Equation~\ref{eqn:erm_distillation}).
We evaluate the models on reasoning and coding benchmarks that benefit from test-time scaling and generation diversity as measured by \passk: GSM8k~\citep{cobbe2021trainingverifierssolvemath}, MATH500~\citep{lightman2023lets}, and MBPP~\citep{austin2021program}. We report \passk{} for $k \in \{1, 4, 8, 16\}$.

We compare models under two settings:
(1) using the sampling temperature that maximizes \passsixteen{}, and
(2) sweeping temperature from 0 to 1.5 (in increments of 0.1) and plotting \passone{} vs. \passsixteen.
This lets us distinguish whether a model is simply stronger overall (higher \passone{} and \passsixteen{}), or whether it has genuinely higher generation diversity—achieving better \passsixteen{} despite similar \passone.

We clarify that in this work we focus on generation diversity as measured by \passk, a standard metric in the LLM reasoning literature~\citep{chen2025rethinkingfinetuningscalingtesttime, dang2025weightensemblingimprovesreasoning,chow2024inferenceawarefinetuningbestofnsampling}. Other notions of diversity, such as creativity~\citep{nagarajan2025rolldicelook} are beyond the scope of this study.

\paragraph{Distilled pretraining unlocks superior  test-time scaling.}
In Figure~\ref{fig:test_time_scaling_main} (top row), we first compare the \passk{} curves for standard pretraining (SPT model) and distilled pretraining (DPT-50 model with 50\% weight of distillation).
We begin by selecting the sampling temperature that maximizes \passsixteen{} performance (a full temperature sweep analysis follows next). 

Observe in Figure~\ref{fig:test_time_scaling_main}(a,b) that while the \dptf model has slighlty worse \passone{} compared to the \spt model, the \dptf model obtains a much higher \passsixteen{} (e.g., 28\% vs. 23\% on GSM). Infact on MATH (Figure~\ref{fig:test_time_scaling_main}b), the \dptf model even starts off worse than \spt on \passk{} at $k=1$, but clearly outperforms it as $k$ increases—exhibiting a \emph{striking crossover phenomenon}. This demonstrates that distilled pretraining yields models with broader coverage and greater diversity in their generations.

\paragraph{Distilled pretraining gives diversity  worth seeing $2\times$ data.}
We now evaluate DPT with an even harder baseline of standard pretraining on 2x data (\sptd, 250B tokens). For DPT, we increase the distillation weight to 90\% (\dptn). As seen in Figure~\ref{fig:test_time_scaling_main} (top row), the \dptn model achieves a better \passsixteen{} than \sptd model across all the three benchmarks—even though \dptn is trained on half the data and has a worse \passone. This highlights the strong diversity gains in model generations from distillation.

In Figure~\ref{fig:test_time_scaling_main} (bottom row), we plot \passone{} vs. \passsixteen{} across temperatures from 0 to 1.5. 
Across all benchmarks—GSM8k, MATH, and MBPP—the \dptn curve consistently lies vertically above the \sptd curve. That is, for any fixed \passone{}, the distilled model achieves a higher \passsixteen. Note that both the models have the same maximum \passone{} (if  one optimizes the temperature for \passone{}), but the distilled model always has a higher maximum \passsixteen{}, or infact a higher \passsixteen{} for any reasonable \passone.
This reinforces that distilled pretraining enables stronger test-time scaling.

\paragraph{Diversity gains even in IsoData setting}
For the results in Figure~\ref{fig:test_time_scaling_main} we used Llama-3.1-8B as the teacher model, which has been trained on more data than what we use to train our 1B models. However, we continue to see better test-time scaling gains even in the IsoData setting where we use a 8B model trained on the same data as a 1B model (1T tokens) as shown in Figure~\ref{fig:isodata_iwl_full} where distilled pretraining (orange curve) outperforms standard pretraining on both GSM8k Pass@16 and MBPP Pass@16.

\paragraph{Higher base model diversity $\rightarrow$ post-training advantages.} The diversity benefits conferred by distillation persist even after post-training on reasoning data, as shown in Figure~\ref{fig:post_train_pass_at_k}(b,c). Again, we observe a crossover-phenomenon, where a model trained with 90\% weight of distillation during pretraining, exhibits lower \passone{} than a 50\% weight counterpart (red vs orange curve in Figure~\ref{fig:post_train_pass_at_k} (b)). However, the model with more distillation heavy pretraining exhibits better test-time scaling due to better diversity in generations.

Finally, in Table~\ref{tab:additional_base_model_evals}, we present evaluations on general language modeling tasks for standard and distillation-pretrained models. As expected, distillation pretraining improves statistical modeling, leading to better performance even on non-reasoning tasks as well, echoing findings in ~\citet{team2024gemma}.

\paragraph{Why does distillation help with diversity?}
When prompts admit multiple plausible continuations—like “I work at”—the ground truth data provides only one answer (e.g., hospital), but a teacher model distributes probability mass across many valid completions (e.g., hospital, gym, cafe). Distillation exposes the student to this richer signal, which intuitively explains why it improves the model’s diversity in it's generations at inference time. We discuss this more formally in the next section.

\begin{tcolorbox}[left=0mm,right=0mm,colback=metablue!8!white,colframe=white]
While \passone{} demands only that the top prediction be correct, \passk{} evaluates whether any of the $k$ outputs are valid—rewarding breadth over precision.
This subtle shift means that correctly ranking one option is not enough; the model must distribute probability mass across multiple plausible answers. Distilled models
excel at this, helping them exhibit better test-time scaling. We will discuss this in more detail in \S~\ref{sec:bayes_optimal_passk}.
\end{tcolorbox}

\section{Building intuition via a bigram sandbox}
\label{sec:bigram}
In the previous section, we saw that distilled pretraining (DPT) rewards models with better for test-time scaling. On the other hand, it seems to impair the in-context learning performance by hurting the learning of induction heads. In this section, we try to dissect the reason behind this tradeoff by analyzing the same in a simple yet powerful sandox of a bigram model.

\subsection{Bigram model: Low-Entropy vs. High-Entropy Rows}
\label{sec:bigram_theory}
To build intuition for our results, consider two illustrative prompts:

\begin{itemize}[left=1pt]
\item \textbf{Low Entropy Prompts: } ``$2+3=$'' with completions: a) $5$, b) $4$, c) $7$ — where a) occurs with probability 1 in natural data.
\item \textbf{High Entropy Prompts: } ``I go to'' with completions: a) office, b) gym, c) restaurant, d) 33 — where a), b), and c) each occur with probability $1/3$ in natural data.
\end{itemize}

\paragraph{Bigram data generation process:}
A bigram model captures a first-order Markov process, where the next token depends only on the current token. Mathematically, it is represented by a matrix $\pi \in \mathbb{R}^{k \times k}$, where each element $\pi_{ij}$ denotes the transition probability from token $i$ to token $j$. 
Our dataset consists of sequences generated from the above bigram model, and the first token is sampled uniformly from the vocabulary.

We categorize each row of the transition matrix $\pi$ as either \emph{low-entropy} or \emph{high-entropy}, based on the entropy of that row relative to a fixed threshold. 
High-entropy rows are akin to prompts that have a diverse completion set (recall ``I go to'' example from above). Low entropy rows then correspond to prompts with less-diverse completions (e.g., ``2+3='').

\begin{figure}[t!]
    \centering
    \begin{subfigure}{0.32\textwidth}
        \centering
        \includegraphics[width=\linewidth]
        {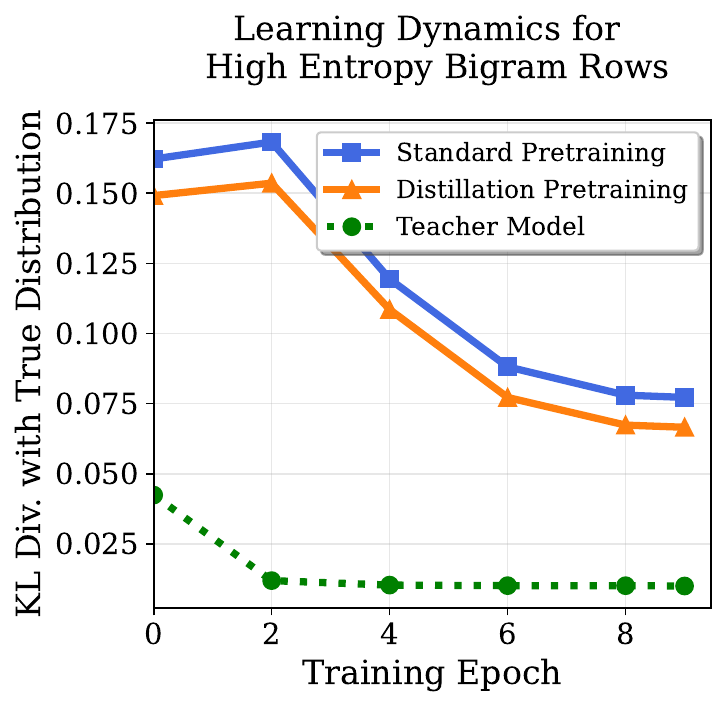}
        \caption{}
    \end{subfigure}
    \hfill
    \begin{subfigure}{0.32\textwidth}
        \centering
        \includegraphics[width=\linewidth]{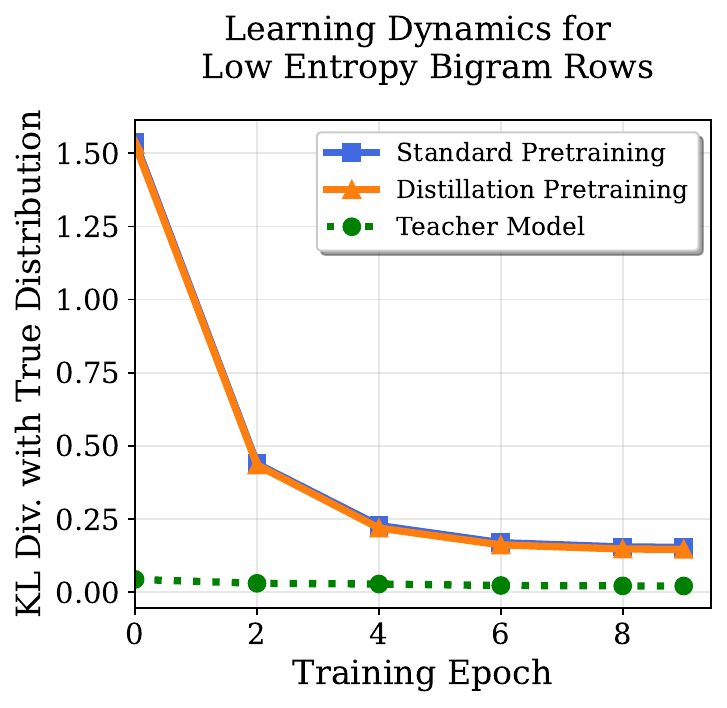}
        \caption{}
    \end{subfigure}
    \hfill
    \begin{subfigure}{0.32\textwidth}
        \centering
        \includegraphics[width=\linewidth]{figures/induction_learning.pdf}
        \caption{}
    \end{subfigure}
    \caption{\textbf{Understanding distillation through the lens of a bigram model (\S~\ref{sec:bigram}):} 
    To dissect why distillation enhances diversity yet impairs in-context learning, we examine these phenomena in a simple yet expressive sandbox—a bigram model~\citep{bietti2023birth, edelman2024evolutionstatisticalinductionheads}. A bigram models a first-order Markov chain represented via a transition probability matrix. \textbf{(a)} We illustrate that distillation particularly aids the learning of high-entropy rows, corresponding to prompts such as ``I work at'', which admit multiple plausible completions (e.g., ``gym'', ``hospital'', ``restaurant''). \textbf{(b, c)} Conversely, distillation offers no advantage for learning low-entropy rows (b), which not only represent deterministic state transitions (prompts), but are also essential for induction head formation as described by~\citet{bietti2023birth}. Moreover, distillation with an imperfect teacher may even slow or hinder learning these low-entropy, induction-head-like rows (c).
}
    \label{fig:bigram_and_bayes_plots}
\end{figure}

\paragraph{Distillation accelerates learning of high-entropy bigram rows}
In Figure~\ref{fig:bigram_and_bayes_plots}(a,b), we present the results of the experiments. The teacher in this case is a bigger model trained on 2 times more data than the student trained from scratch (further details on the experimental setup are in the Appendix~\ref{app:bigram_experiment_details}). As we can see, models trained from scratch and models trained via distillation are both at par when it comes to the low entropy rows (Figure~\ref{fig:bigram_and_bayes_plots}b). A real distinction appears in how well they approximate the high entropy rows, where the distilled model performs better, i.e., it requires fewer samples to achieve a better approximation of the high-entropy row(Figure~\ref{fig:bigram_and_bayes_plots}a). We now formalize the intuitions behind the above arguments.

\paragraph{Sample complexity analysis for bigram model} Each row of the bigram matrix $\pi$ is $p$-sparse, i.e., contains at most $p$ non-zero entries. We consider sequences of length two. Both the scratch-trained and distilled student models are parameterized by bigram matrices $\pi^{\mathsf{scratch}}$ and $\pi^{\mathsf{distill}}$, respectively, while the teacher is parameterized by $\pi^{\mathsf{teacher}}$. Here, $\pi^{\mathsf{scratch}}$ and $\pi^{\mathsf{distill}}$ are solutions to \eqref{eqn:erm_distillation} with $\alpha = 0$ and $\alpha = 1$, respectively.

\begin{tcolorbox}[left=0mm,right=0mm,colback=metablue!8!white,colframe=white]
\begin{proposition}(informal) In bigram learning with $p$-sparse rows \hfill
\begin{itemize}[left=1pt]
\item Sample complexity when training with distillation is $ \mathcal{S}_{\text{distill}}= \mathcal{O}(k\log k$).

\item Sample complexity when learning bigram without distillation is $\mathcal{S}_{\text{standard}} \approx \frac{p}{\epsilon^2}\mathcal{S}_{\text{distill}}$, where $\epsilon$ is the upper bound on the the approximation error.

\end{itemize}
\end{proposition}
\end{tcolorbox}

Consider first the high-entropy setting where the row sparsity $p = \mathcal{O}(k)$, where $k$ denotes the vocab size. The standard pretrained model requires $\mathcal{O}(k^2 \log k)$ samples, whereas the distilled model needs only $\mathcal{O}(k \log k)$. In contrast, in the low-entropy setting where $p$ is constant, both models have sample complexity at most $\mathcal{O}(k \log k)$. This reflects the empirical observations from Figure~\ref{fig:bigram_and_bayes_plots}(a,b) where we observe distillation accelerating the learning of high-entropy rows but no difference for low-entropy rows.

\textit{Proof sketch.} To prove the first part, note that once a token $i$ is observed, the teacher provides perfect supervision via the soft label $\pi^{\mathsf{teacher}}_i$, leading to $\pi^{\mathsf{distill}}_i = \pi^{\mathsf{teacher}}_i$. Hence, it suffices to observe each token at least once. By standard coupon collector tail bounds, $\mathcal{O}(k \log k + k \log (1/\delta))$ samples ensure this with probability at least $1 - \delta$.

To prove the second part, we show that observing token $i$ in the first position at least $\mathcal{O}(p/\epsilon^2)$ times ensures $\mathbb{E}[|\pi^{\mathsf{scratch}}_i - \pi_i\|_1] \leq \epsilon$. Thus, each row must be seen $\mathcal{O}(p/\epsilon^2)$ times. Applying tail bounds from the generalized coupon collector’s problem (with $m$ observations per coupon) yields the desired sample complexity.

\subsection{Why does induction head learning slow down for distilled models?}
\label{sec:induction_head_detail}

Recall from \S~\ref{sec:icl} that distillation impairs the learning of induction heads—key circuits for in-context learning. We now revisit this phenomenon by detailing the induction head learning setup in our bigram sandbox.

Following \citet{bietti2023birth}, we modify the bigram model to embed an \emph{induction‑style pattern} using \textbf{trigger tokens}. A trigger token is a special token in the vocabulary such that whenever it appears, it is always followed by a fixed token within that sequence. Importantly, this fixed token is different for different sequences but remains the same for all trigger occurrences within a single sequence.

Formally, before generating each sequence, we randomly choose a “copy target” token $c \in \{1, \dots, k\}$. We then alter the bigram transition matrix $\pi$ so that whenever the current token is the trigger (denoted $i=t$), the next token is deterministically $c$. Mathematically:

$$
\tilde{\pi}_{ji} = 
\begin{cases}
\pi_{ji} & \text{if } i \ne t \\
\mathsf{I}(j = c) & \text{if } i = t
\end{cases}
$$

Sampling from $\tilde{\pi}$ produces a setting where the optimal strategy is to learn to \emph{copy} the token ($c$) following a trigger token (the token $t$ in the above case)—mimicking the behavior of induction heads in real LLMs~\citep{olsson2022incontextlearninginductionheads, bietti2023birth}.

The difference between standard pretraining and distillation emerges in the supervision signal.
\begin{itemize}[left=1pt]
\item In standard pretraining, encountering a trigger yields a one‑hot ground‑truth label for the next token—clean and unambiguous supervision.
\item In distillation with a \emph{perfect} teacher, the soft label distribution is also exactly one‑hot, so the supervision is identical. In practice, however, teachers are imperfect: they may assign non‑zero probability mass to distractor tokens. This produces a slightly higher‑entropy target distribution, effectively injecting noise into what should be a deterministic mapping.
\end{itemize}

This explains the consistent drop in induction‑head task performance for distilled models, as observed in Figure~\ref{fig:title_base_model_passk}(c).

\subsection{Why does \texorpdfstring{\passk{}}{\passk{}} improve for distilled models?}
\label{sec:bayes_optimal_passk}

\paragraph{Demistifying \passk{} trends:}

In Figure~\ref{fig:title_base_model_passk}(b), we saw a puzzling finding. The distilled model can start with a worse \passone{} and can have a much better \passk. Is this a mere accident, or does there exist a deeper principle behind the observations?

Suppose that our data consists of one fixed prompt $x$, which is followed by three options $y=\{0,1,2\}$. The true probabilities are $p(y=0|x)=\frac{1}{2}+\epsilon$, $p(y=1|x)=\frac{1}{2}-\epsilon$ and $0$ with $\epsilon>0$. Define three classifiers: 

\begin{itemize}[left=1pt ]
    \item \textbf{Bayes optimal classifier, C1:} Assigns a probability $1$ to class $0$ and achieves the optimal \passone{} accuracy of $\frac{1}{2}+\epsilon$.
    \item \textbf{Diverse classifier with right coverage, C2:} Assigns a probability of $\frac{1}{2}$ to both classes $0$ and $1$. This classifier achieves a suboptimal \passone{} accuracy of $\frac{1}{2}$. 
    \item \textbf{Diverse classifier with wrong coverage, C3:} Assigns a probability of $\frac{1}{2}$ to classes $0$ and $2$. This classifier achieves a suboptimal \passone{} accuracy of $\frac{1}{4}+ \frac{\epsilon}{2}$. 
\end{itemize}

\begin{wrapfigure}[20]{r}{0.35\textwidth} %
    \vspace{-2em}
    \centering
    \includegraphics[width=\linewidth]{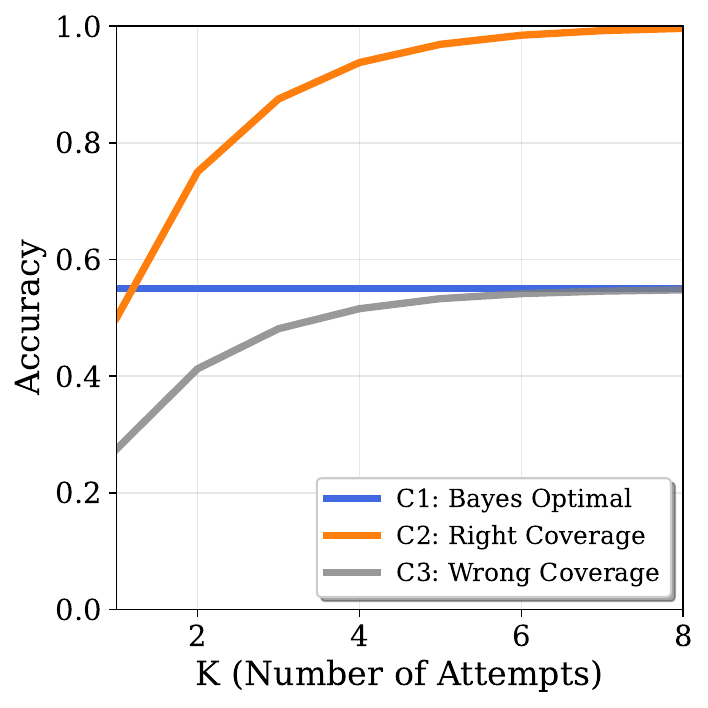} %
    \caption{\textbf{Bayes optimal for \passone{} is not optimal for \passk.} A diverse classifier with correct coverage (C2) outperforms the Bayes optimal classifier (C1) at higher $k$, while incorrect coverage (C3) remains suboptimal. Coverage—not just \passone{}—is key to improving \passk.}
    \label{fig:bayes_optimal}
    \vspace{-2em}
\end{wrapfigure}

Interestingly, observe that the \passk{} accuracy of \textbf{C1} is $\frac{1}{2}+\epsilon$ for all $k$. The \passk{} accuracy of \textbf{C2} is $1- (\frac{1}{2})^k$ for all $k$. The \passk{} accuracy of \textbf{C3} is $(\frac{1}{2}+\epsilon)(1- (\frac{1}{2})^k)$ for all $k$. As shown in Figure~\ref{fig:bayes_optimal}, the classifier \textbf{C2} exhibits crossover over the Bayes optimal classifier \textbf{C1}. Thus, the Bayes optimal classifier is suboptimal at higher \passk. Further, \textbf{C3}'s support does not contain the support of the true distribution, highlighting the importance of right coverage over the correct solution space.

The above example leaves us with the question that if the Bayes optimal classifier is not optimal for \passk{}, then what is? We derive this classifier below.

\paragraph{Generalized Bayes optimality for \passk{}} In this section, we restrict ourselves to binary classification tasks with the true probability distribution over labels $y$ conditional on $x$ denoted as $p(y|x)$. 

Recall the definition of a Bayes optimal classifier for binary classification. For each $x$ in the support of the training distribution, the classification rule is 

\begin{equation}
    \begin{cases}
    0  \; \; \; \;\;\;\; p(y=0|x)> \frac{1}{2} \\ 
    1 \; \; \; \;\;\;\; p(y=1|x)\leq  \frac{1}{2}. 
    \end{cases}
\end{equation}

Define a general classifier which assigns a probability $\alpha(x)$ to class $1$ and $\beta(x)$ to class $0$. 

\begin{tcolorbox}[left=0mm,right=0mm,colback=metablue!8!white,colframe=white]
\begin{theorem}
\label{thm:bayes_optimal_passk}
The generalized Bayes optimal classifier that achieves the optimal \passk{} assigns for each $x$ in the training distribution
\begin{equation}
\label{eq:thm_1}
    \alpha^{*}(x) = \frac{\Big(\frac{p(y=1|x)}{p(y=0|x)}\Big)^{\frac{1}{k-1}}}{1+\Big(\frac{p(y=1|x)}{p(y=0|x)}\Big)^{\frac{1}{k-1}}}.
\end{equation}
\end{theorem}
\end{tcolorbox}

\begin{proof}

\passk{} accuracy of a classifier checks if at least one of the $k$ attempts of the classifier predicts the label correctly. 
For a fixed $x$ in the support of the training distribution, the \passk{} accuracy of this classifier is stated as 

\begin{equation}
    p(y=1|x)\big(1-(\beta(x))^k\big) + p(y=0|x)(1-(\alpha(x))^k).
\end{equation}

To understand the above expression, let us look at the first term. Conditional on $y=1,x$, $\big(1-(\beta(x))^k\big)$ is the probability that at least one of the attempts by the model says class 1. 

To simplify notation, let us write $p(y=1|x)$ as $p$, $\alpha(x)$ as $\alpha$ and rewrite the above as 
\begin{equation}
    p\big(1-(1-\alpha)^k\big) + (1-p)(1-\alpha^k) .
\end{equation}

The function is concave in $\alpha$ for $\alpha \in [0,1]$ and $k \geq 1$, with second derivative given by $-(k)(k-1)\big(p(1-\alpha)^{k-2}+(1-p)\alpha^{k-2}\big)$.
Setting the first derivative to zero gives 

$$\alpha^{*} = \frac{\Big(\frac{p}{1-p}\Big)^{\frac{1}{k-1}}}{1+\Big(\frac{p}{1-p}\Big)^{\frac{1}{k-1}}}.$$

Thus, the generalized Bayes optimal classifier is as given in Eq.~\ref{eq:thm_1}.
Observe that as $k$ approaches 1 from the right, the expression reduces to the standard Bayes optimal classifier: if $p(y=1|x) > 1/2$, then $\alpha^{*}(x) = 1$; otherwise, $\alpha^{*}(x) = 0$. 
This completes the proof.

\end{proof}

A few key remarks follow. For $k=1$, the Bayes optimal classifier is $\alpha^*(x) = \mathsf{I}\big(p(y=1|x) > \frac{1}{2}\big)$. Optimal \passone{} requires only correct ordering of class probabilities—not precise estimates of $p(y=1|x)$. In contrast, optimal \passk{} demands accurate estimation of $p(y=1|x)$. Distilled models better approximate these distributions, especially in high-entropy settings. While this may not improve \passone{}, it yields superior \passk{} performance.

\section{Practitioners Guidelines}
\subsection{Token Routing: Mitigating the Drop in In-Context Learning}
\label{sec:token_routing}

\begin{figure}[t!]
    \centering
    \begin{subfigure}{0.235\textwidth}
        \centering
        \includegraphics[width=\linewidth]{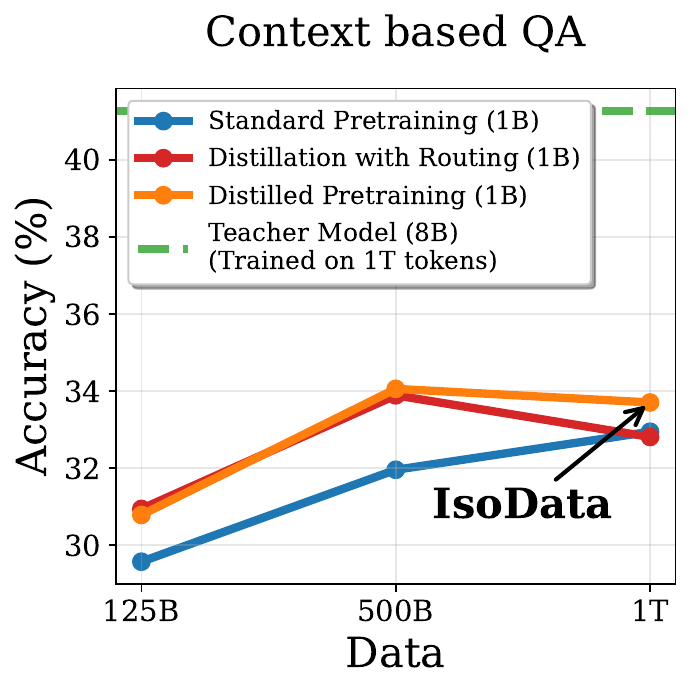}
    \end{subfigure}
    \hfill
    \begin{subfigure}{0.245\textwidth}
        \centering
        \includegraphics[width=\linewidth]{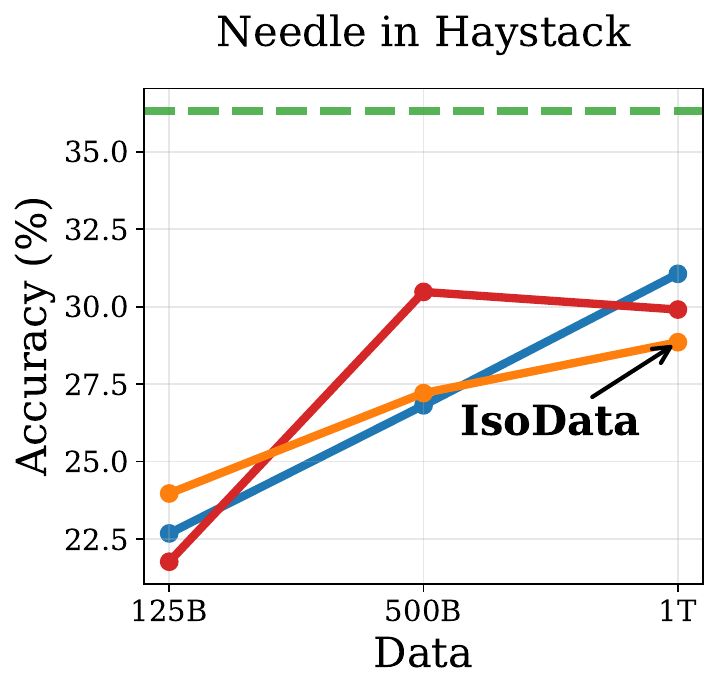}
    \end{subfigure}
    \hfill
    \hfill
    \begin{subfigure}{0.235\textwidth}
        \centering
        \includegraphics[width=\linewidth]{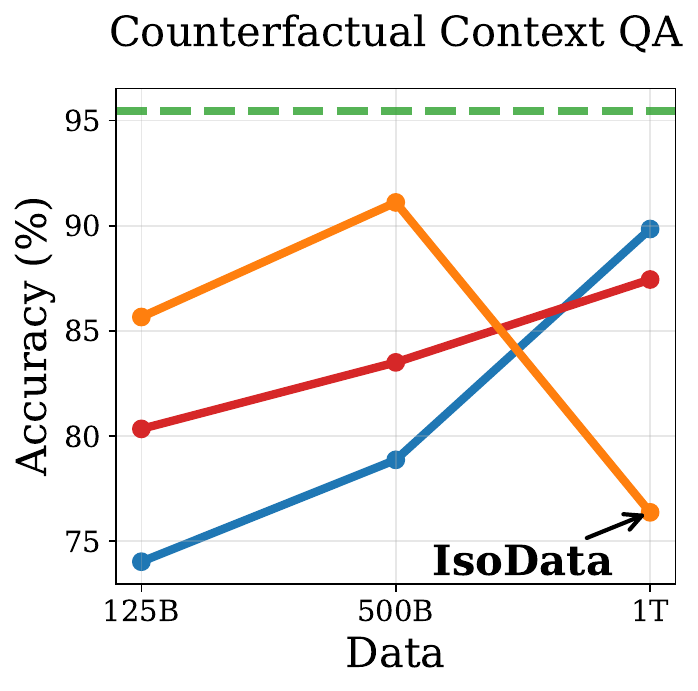}
    \end{subfigure}
    \hfill
    \begin{subfigure}{0.245\textwidth}
        \centering
        \includegraphics[width=\linewidth]{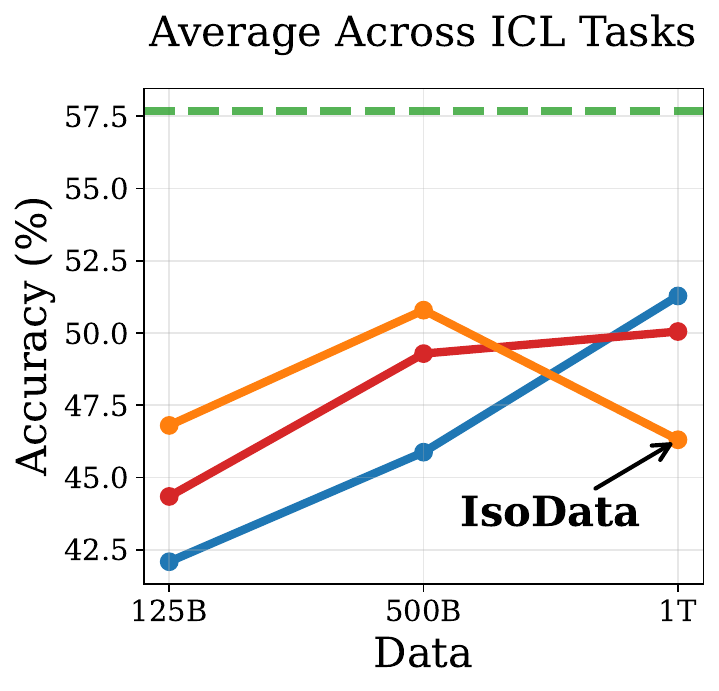}
    \end{subfigure}
    \caption{\textbf{Token Routing: Mitigating the Drop in In-Context Learning(\S~\ref{sec:token_routing}): } Distilled models often struggle on ICL tasks due to softening of supervision on low-entropy (near-deterministic) tokens—critical for copying behavior via induction heads. To mitigate this, we apply token routing: for each input, we skip the distillation loss on the 15\% lowest-entropy tokens, using only ground-truth supervision there.
    This strategy (red curve) improves over vanilla distillation (orange) on two of three ICL tasks, partially closing the gap to standard pretraining (blue). As shown in Table~\ref{tab:entropy_routing_additional_evals}, these gains come without hurting standard language modeling performance.
    } 
    \label{fig:entropy_routing_15p}
\end{figure}

In Section~\ref{sec:icl}, we saw that distillation performs poorly on in-context learning (ICL) tasks compared to standard pretrained models—especially when the teacher and student are trained on the same amount of data.
Recall that this is because induction tasks are built on low-entropy mappings where distillation doesn't help.

To mitigate this, we propose a simple yet effective strategy: token routing. Recall from Equation~\ref{eqn:erm_distillation} that during distilled pretraining, there are two terms in the loss-one for loss with ground truth labels and the other with teacher's label (distillation loss term).
Rather than applying distillation loss with the teacher's label on all tokens, we dynamically adjust the supervision based on the entropy of the teacher’s output. Specifically, given an input sequence, we first compute the teacher's soft labels for the sequence. We then drop the distillation loss term for $x\%$ of the positions with lowest entropy in teacher's label—falling back to only the standard hard-label supervision with the ground truth here.

In Figure~\ref{fig:entropy_routing_15p} we share results when routing $x=15\%$ of the tokens. In two of three tasks, Needle in a Haystack and Counterfactual Context-Based QA, token routing leads to noticeable improvements over vanilla-distilled pretraining, partially closing the performance gap with standard pretraining. On Context-based QA, no performance gain is expected because vanilla distillation performs better than standard pretraining. Moreover, this does not hurt the performance on standard language modeling tasks (Table~\ref{tab:entropy_routing_additional_evals}). Infact, this also reinforces the intuition that gains from distillation come from high-entropy teacher labels.
In Appendix~\ref{app:token_routing} we share results when routing $x=30\%$ of the tokens, which does not help further on ICL tasks while also hurting standard benchmark performance.

While preliminary, this demonstrates how token-level curation can adapt distillation to better suit modern LLM centric objectives like in-context learning. We hope our work motivates future research in developing more nuanced strategies.

\subsection{NTP vs. MTP vs. Distillation: Which Yields Better Diversity?}
\label{sec:ntp_vs_mtp_vs_distill}
\begin{figure}[t!]
    \centering
    \begin{subfigure}{0.32\textwidth}
        \centering
        \includegraphics[width=\linewidth]{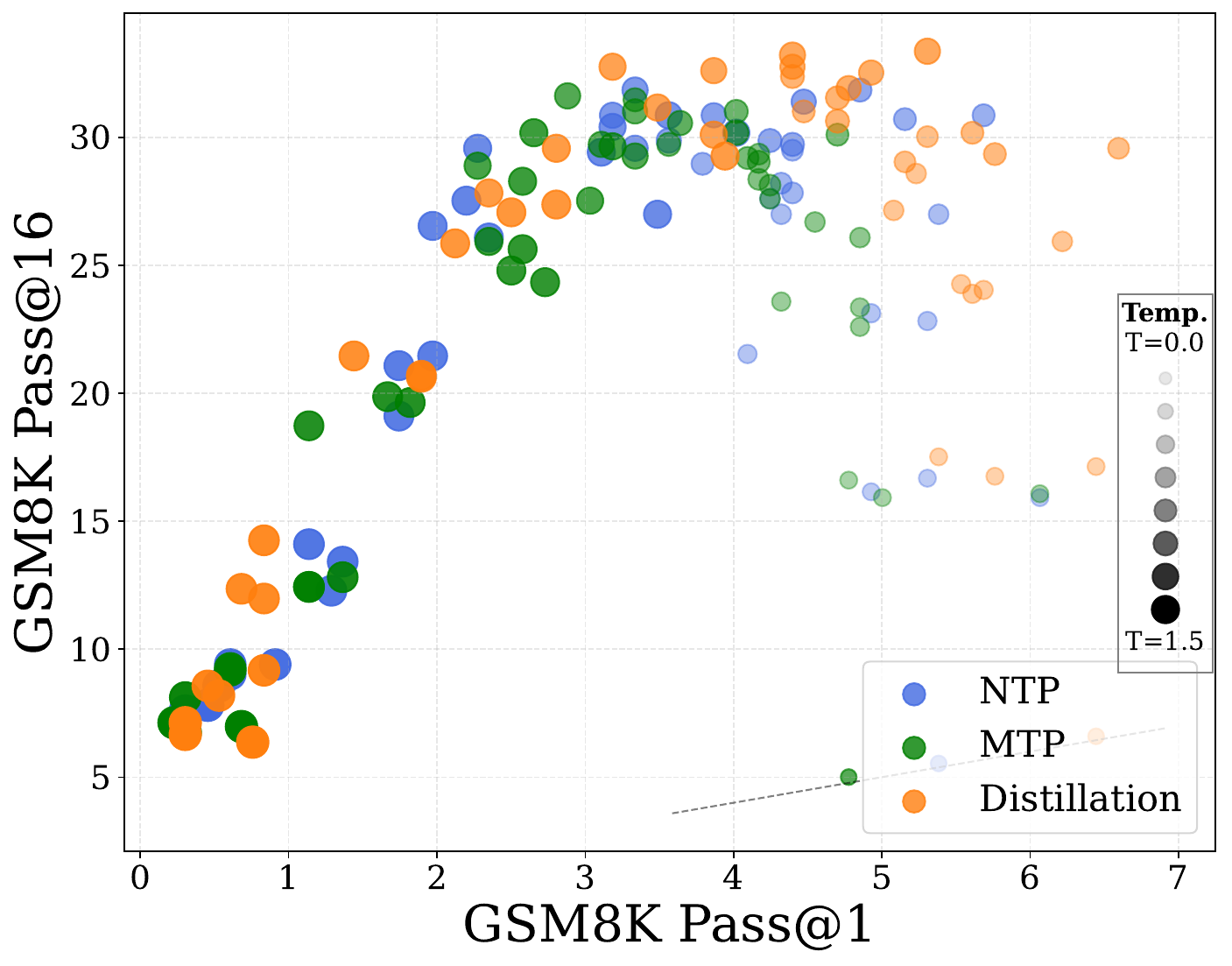}
    \end{subfigure}
    \hfill
    \begin{subfigure}{0.32\textwidth}
        \centering
        \includegraphics[width=\linewidth]{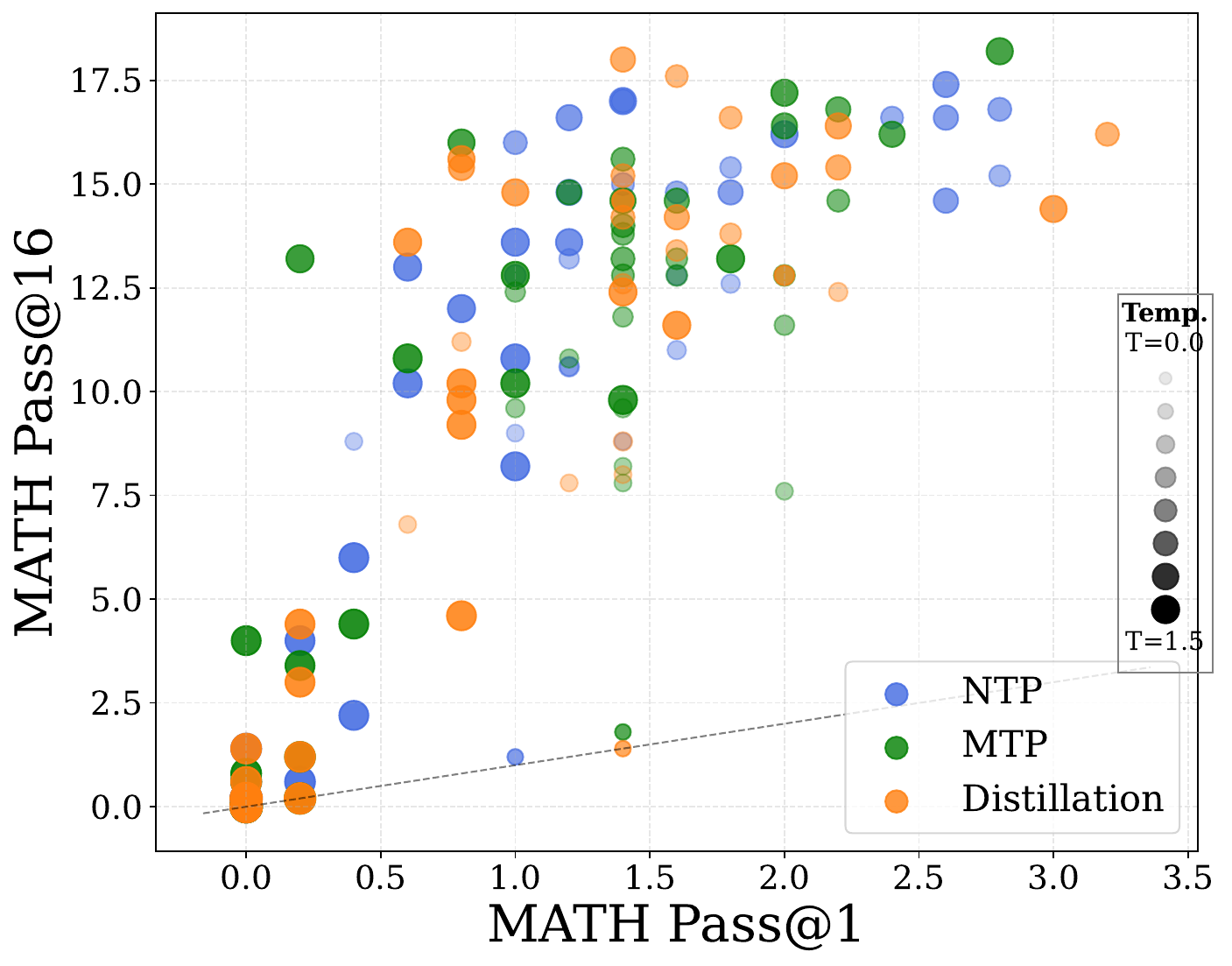}
    \end{subfigure}
    \hfill
    \begin{subfigure}{0.32\textwidth}
        \centering
        \includegraphics[width=\linewidth]{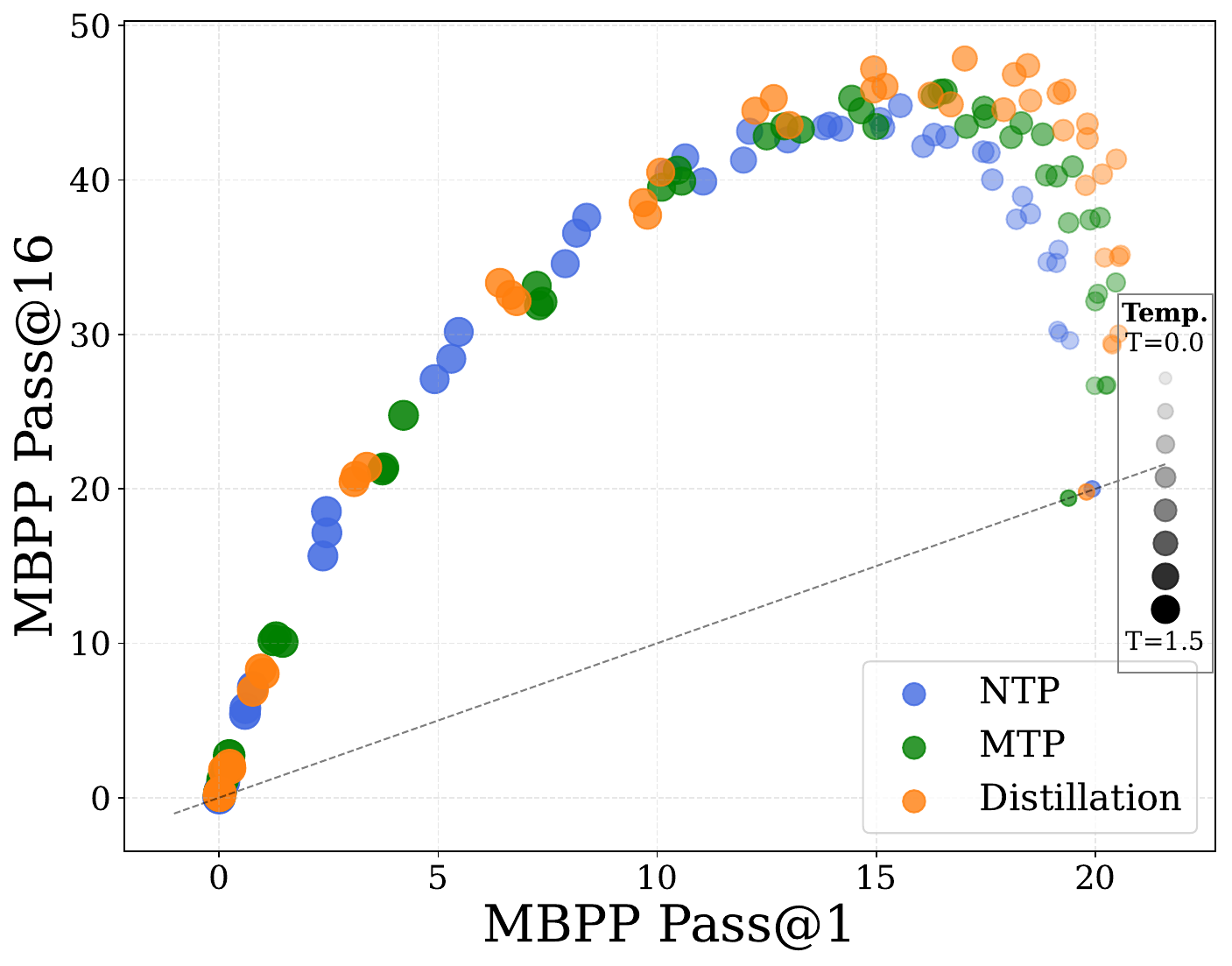}
    \end{subfigure}
    \caption{\textbf{NTP vs MTP vs Distillation: }We compare 1B models trained on 1T tokens via (1) standard next‑token prediction (NTP), (2) multi‑token prediction (MTP), and (3) distillation from an 8B teacher trained on the same 1T tokens (\emph{IsoData} setting). We plot \passone{} vs \passsixteen{} curve. Distillation curve lies generally above MTP on GSM8k and MBPP, and matches it on MATH—despite no data advantage. In real‑world setups, where teachers have seen more data, the gains from distillation are expected to be even larger.
        } 
    \label{fig:ntp_vs_mtp_vs_distill}
\end{figure}

In this work, we showed that distillation produces models particularly well-suited for test‑time scaling—primarily due to their richer generation diversity. In parallel, recent works on multi‑token prediction (MTP)~\citep{gloeckle2024betterfasterlarge} have also emerged as a promising way to train inherently diverse models~\citep{nagarajan2025rolldicelook}. This raises a natural question for practitioners: given the choice, should one invest in MTP or in distillation?

To answer this, we compare three pretraining strategies for 1B models:
(1) standard next-token pretraining (NTP),
(2) MTP, and
(3) distillation from an 8B teacher trained on 1T tokens same as the student corpus.

In Figure~\ref{fig:ntp_vs_mtp_vs_distill} we plot \passone{} vs \passsixteen{} for the three pretraining choices. We observe that the curve for distilled pretraining lies above those of MTP and NTP. This implies that given any reasonable \passone{}, distilled model exhibits higher \passsixteen{} (on GSM8k and MBPP) or similar \passsixteen{} (on MATH)  compared to multi-token pretraining.
This is notable given the fairness of our setup—using a teacher trained on exactly the same data as the student. In practice, where teachers are often stronger because they have seen more data, the advantage of distillation is likely to be even greater. These findings reinforce distillation’s strong value proposition for practitioners aiming to train small models that excel under verifier‑driven inference settings~\citep{alphaevolve,snell2024scalingllmtesttimecompute}.

\subsection{Base vs. RL model: What makes a better teacher?}
\label{sec:base_vs_ins_vs_rl}
\begin{figure}[t!]
  \centering
  \includegraphics[width=\textwidth]{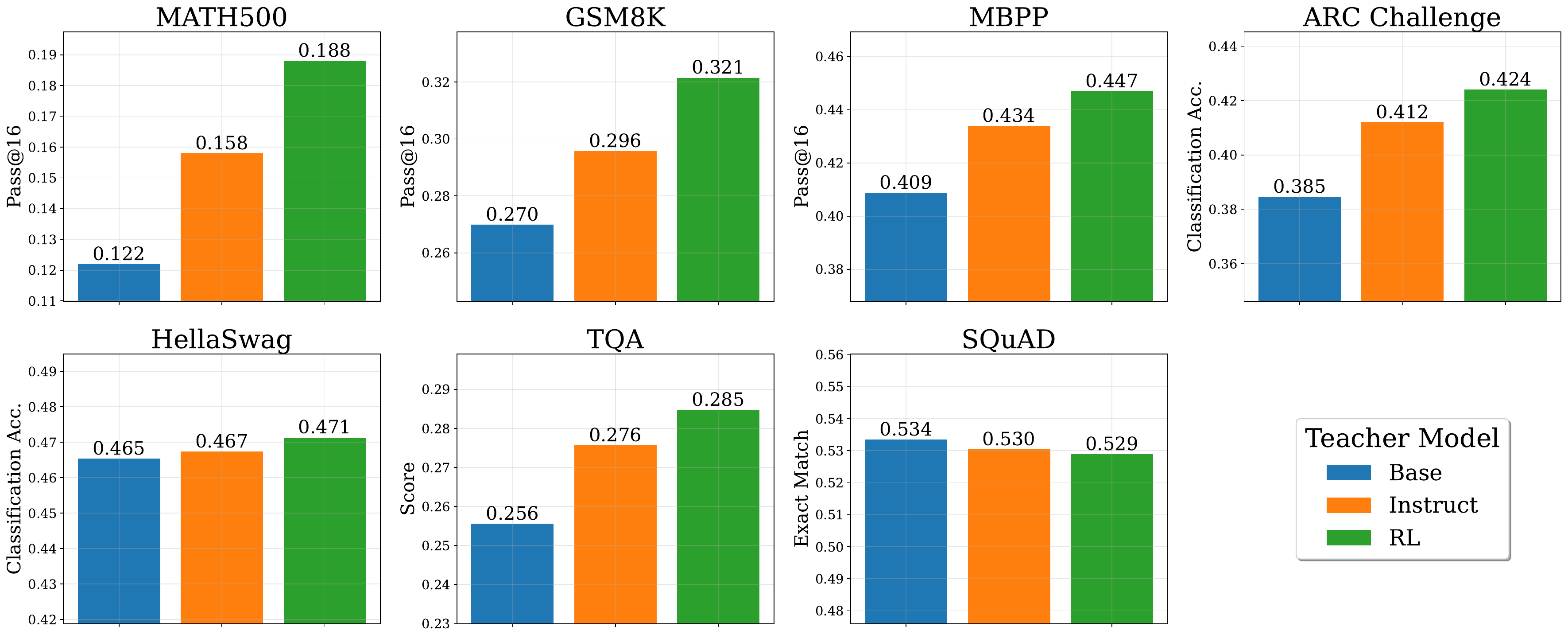}
\caption{\textbf{What makes a better teacher: Base vs Instruct vs RL model(\S~\ref{sec:base_vs_ins_vs_rl}): }We compare 1B student models distilled from three version of a model: base, instruction-tuned, and RL-trained. The RL-trained teacher consistently yields the best student—across reasoning (MATH500, GSM8k), coding (MBPP), and even general benchmarks (TQA, HellaSwag, ARC). This suggests that stronger teacher performance may outweigh alignment mismatches with the pretraining objective. Despite common practice favoring base models as teachers (e.g., Gemma, Llama-3.2), our findings highlight the potential of RL-trained models as superior teachers for distilled pretraining.}
  \label{fig:base_vs_instruct_vs_rl}
\end{figure}

A general question we had while distilling with a teacher was—what version of the teacher model should be used: the base version, the instruction-tuned version, or the RL-trained version?

At first glance, the base model appears to be the better choice—it aligns more naturally with the pretraining objective of free-form sentence completion and also with the current practice~\citep{team2024gemma,meta2024llama32}. In contrast, instruction-tuned and RL-trained models are more tailored to QA-style prompting, making them less aligned with the standard pretraining setup. But on the other hand, the Instruct and RL versions are often better in many capabilities and performance on downstream benchmarks, particularly for reasoning and code tasks. At the same time, recent works like ~\citet{dang2025weightensemblingimprovesreasoning} highlight that Instruct and RL models suffer from reduced diversity in their generations, which suggests they might not be the better choice as a teacher during pretraining.

We try to answer this puzzle empirically by training student models of 1B size, distilled from 3 versions of a 8B teacher model: the base Llama-3.1-8B, its instruction-tuned counterpart, and the RL-trained variant optimized for reasoning. 
Interestingly, the results in Figure~\ref{fig:base_vs_instruct_vs_rl} favor the Instruct and the RL-trained teacher—across the board. The student distilled from RL trained teacher not just outperforms on reasoning and coding benchmarks (which might be expected), but also on general language modeling tasks like HellaSwag and TQA. This finding indeed surprised us as well. Note that many distillation pretrained models currently like Gemma series~\citep{team2024gemma,team2025gemma} and the Llama-3.2 series~\citep{meta2024llama32} are distilled using base version of a large model as the teacher. Infact, even in our work we used base model as the teacher. We hope these insights help inform better teacher choices for future distilled pretraining.

\subsection{Top-k sampling distillation}
\label{sec:topk}

Rather than using the teacher's soft distribution over the whole vocabulary, a common practice~\citep{team2025gemma} is to sample $k$ logits per token based on the teacher's original output distribution, and then re-normalize the weights of the sampled logits to get a sparse label(logits not samples are set to 0). This reduces the cost of distillation. In this section, we try to answer whether the choice of $k$ here has (if) any impact on downstream performance. We note here that the case of $k=1$ interestingly corresponds to standard pretraining with a ``token level synthetic data'' from the teacher model. 

\begin{figure}[t!]
  \centering
  \includegraphics[width=\textwidth]{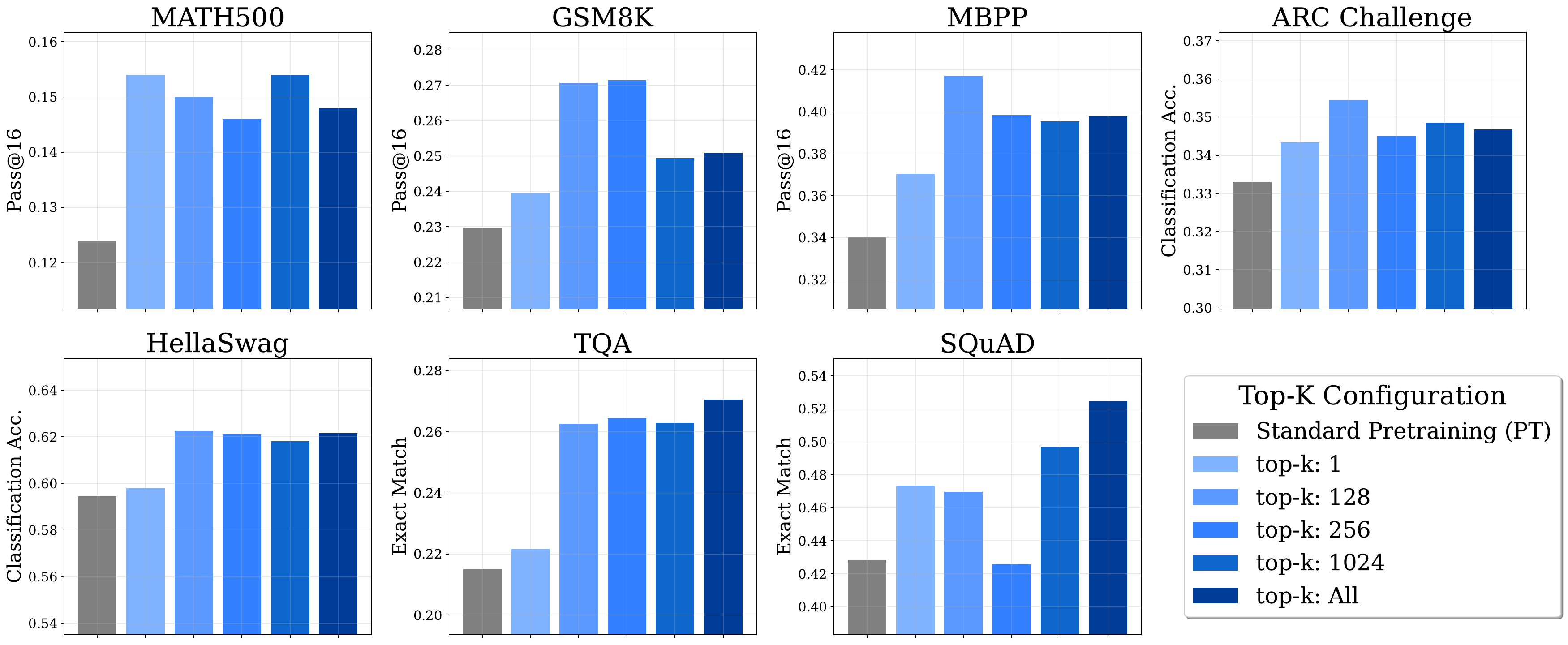}
\caption{\textbf{Top-k sampling distillation(\S~\ref{sec:topk}): }We compare using sparse soft target label by sampling k-logits per token. $k=1$ corresponds to a token level synthetic data albeit without any soft labels, and outperforms standard pretraining. Using richer soft labels ($k=$ 128, 256, 1024, or All) further improves performance, but no clear winner emerges among them.}
  \label{fig:topk}
\end{figure}

Figure~\ref{fig:topk} shows the results. We observe two clear trends: (1) Even top-$k=1$ outperforms standard pretraining, likely due to the use of synthetic tokens and the teacher filtering out outlier tokens from the ground truth; and (2) using $k \in {128, 256, 1024, \text{All}}$ leads to better performance than top-$k=1$, as the benefits of soft label distributions begin to take effect. However, there is no consistent trend indicating which $k$ (other than $k=1$) performs best.

\section{Related works}

\paragraph{Classical paradigm of distillation}

The story of distillation begins with \cite{buciluǎ2006model}, where the technique was introduced to compress an ensemble of models into a single model. Subsequently, \cite{ba2014deep} proposed a form of distillation wherein a student is trained by minimizing a regression loss against teacher logits. Later, \cite{hinton2015distilling} introduced the most prominent form, combining ground-truth labels with soft labels from a teacher model. Distillation further evolved into various forms: self-distillation \cite{furlanello2018born}, where earlier student checkpoints act as teachers; progressive distillation \citep{mirzadeh2020improved}, in which earlier teacher checkpoints progressively guide the student; and generalized distillation \cite{lopez2015unifying}, which integrates standard distillation with the privileged information framework.

An extensive theoretical literature has examined distillation through multiple lenses. For instance, \cite{phuong2019towards, safaryan2023knowledge} adopted an optimization perspective to explain distillation's benefits, while \cite{menon2021statistical} considered the sample complexity perspective. Given the vast breadth and depth of research on distillation, we refer the reader to \cite{gou2021knowledge} for a comprehensive overview.

\paragraph{Modern paradigm of distillation}
In the past year, we've witnessed a resurgence of distillation in the context of modern LLMs. Both the Llama-3.2 (1B and 3B models)~\citep{meta2024llama32} and Gemma model families (sizes ranging from 3B to 27B) \citep{team2024gemma, team2025gemma} rely heavily on pretraining distillation mechanisms. These models primarily employ the prominent weighted loss introduced by \cite{hinton2015distilling}. Additionally, synthetic data, generated by teacher models, is now commonly used to enrich pretraining corpora, effectively constituting another form of hard-label distillation. \cite{cha2025knowledge} analyzed distillation using synthetic data generation (hard-label distillation), where students learn from samples drawn directly from the teacher model. In contrast, our analysis focuses on \cite{hinton2015distilling}-style pretraining distillation, where students learn from soft labels provided by the teacher, leading us to distinct conclusions from \cite{cha2025knowledge} regarding prediction diversity. We believe that this happens because of the difficulty in sampling diverse synthetic pretraining data (hard labels) from the teacher. Recently, ~\citet{busbridge2025distillationscalinglaws} discuss how distillation might not be helpful under certain compute-matched settings. However, in \S~\ref{sec:isodata} we argue that incorporating teacher logit computation cost might not be the correct setting, and it is more important to consider data-constrained settings.

\citet{li2021dynamicknowledgedistillationpretrained} discuss using small teacher models for tokens where the student model predictions are less confident and vice versa. ~\citet{cho2019efficacyknowledgedistillation,zhang2024lawcapacitygapdistilling, mirzadeh2019improvedknowledgedistillationteacher, zhang2023liftingcursecapacitygap,beyer2022knowledgedistillationgoodteacher} highlight bigger teacher is not always better and propose various ways to mitigate capacity mismatch between student and the teacher to improve distillation.

Beyond pretraining, distillation is increasingly used in post-training. For example, DeepSeek R1 released distilled models via off-policy distillation, where students are fine-tuned on teacher-generated traces~\citep{muennighoff2025s1simpletesttimescaling}. In contrast, on-policy distillation~\citep{agarwal2024onpolicydistillationlanguagemodels,yang2025qwen3technicalreport} uses student-generated traces with logit supervision from the teacher, and has been shown to outperform off-policy methods. In this work, we study logit distillation during pretraining (while using the ground truth data) and highlight the distinct trends and tradeoff's which emerge compared to standard pretraining.

\paragraph{Diversity for test-time search in LLMs}
Diversity in generations is crucial for test-time scaling of LLMs. This is an especially required for open-ended discovery and reasoning tasks, where verification of the correct answer is easy, thus multiple attempts can be done at a problem. ~\citep{alphaevolve, setlur2024rewardingprogressscalingautomated, lifshitz2025multiagentverificationscalingtesttime, beeching2024scalingtesttimecompute}.
In fact, a long line of work focuses on explicitly improving the diversity of generations in LLMs at inference time via diversity aware finetuning~\citep{sessa2024bondaligningllmsbestofn, zhang2024forcingdiffusedistributionslanguage, chow2024inferenceawarefinetuningbestofnsampling, chen2025rethinkingfinetuningscalingtesttime}.  Another line of work explores inference time decoding strategies~\citep{chen2024alphamathzeroprocesssupervision} for promoting diversity if generations and hence better test-time scaling.
While all these works focus on patch-fixing the diversity issue via model finetuning, we highlight an intriguing albeit intuitive gain in diversity of base model itself when pretraining with distillation. This is of even more importance given recent findings that post-training or RL simply sharpens base model distribution. ~\citet{yue2025doesreinforcementlearningreally} shows that base model is better than RL trained model on \passk{} for high $k$. Having a base model with high diversity is also crucial for effective post-training with reinforcement learning via verifiable reward (RLVR), as discussed in ~\citet{dang2025weightensemblingimprovesreasoning}.

\section{Discussion and Concluding Remarks}
While distilled pretraining was notably absent in early LLM training pipelines, it has recently regained prominence, as exemplified in Gemma and Llama series (3.2 and Maverick) which rely solely on distilled pretraining.

In this work, we first addressed a common question arising from the renewed interest in distilled pretraining: Is distillation simply a proxy for accessing the extensive data seen by a larger teacher model, or will it offer inherent benefits even if the student model is trained on all the dataset as seen by the teacher? This question is even more important given the  data constrained regime for modern LLMs. Our findings affirmatively demonstrate that the value of distillation extends beyond mere data augmentation. Specifically, distilled pretraining naturally produces models exhibiting greater generation diversity, inherently enhancing test-time scaling capabilities. This insight is especially significant given recent evidence suggesting that post-training and reinforcement learning methods primarily just sharpen existing base model distributions, with base models often matching post-trained models in higher \passk{} scenarios~\citep{yue2025doesreinforcementlearningreally}. Distillation thus provides a foundational improvement via pushing the base model performance itself rather than a post-hoc fix.

With modern LLMs hitting the data wall and growing interest in enhancing capabilities for open-ended discovery and reasoning tasks, our findings are both timely and impactful. An immediate next step would be to tailor, integrate and evalaute distilled pretraining with other recent advances in pretraining like multi-token pretraining~\citep{gloeckle2024betterfasterlarge,nagarajan2025rolldicelook} and future-aware pretraining~\citep{thankaraj2025lookingtoken,gerontopoulos2025multitokenpredictionneedsregisters} for improving diversity of base models.

In our study, we proposed applying distillation selectively on a subset of tokens—particularly to mitigate cases where full-token distillation may hurt performance. More broadly, current pretraining datasets have largely been curated from common crawl with standard next-token pretraining paradigms in mind. Moving forward, a highly promising research direction would be the development of pretraining datasets and curation approaches specifically optimized for distilled pre-training.

Moreover, given the widespread adoption of distillation in post-training phases—such as fine-tuning on reasoning traces generated by larger models—another intriguing avenue is to investigate whether using the same teacher model for both pretraining and post-training distillation could better align these two phases. Our work provides preliminary insights into several practical design choices practitioners face during distilled pretraining, and we hope these contributions support the community in advancing this promising line of research.

\section{Acknowledgments}
The authors thank Divyat Mahajan for his help with the initial setup of the codebase and infrastructure. The authors also gratefully acknowledge the helpful discussions with Badr Youbi Idrissi, Mohammad Pezeshki, Mathurin Videau, Sharut Gupta, Sarthak Mittal, Andrei Nicolicioiu and Julia Kempe. We really appreciate the detailed feedback on the initial drafts from Vaishnavh Nagarajan and 
Christina Baek. We thank Jacob Springer for the helpful discussion on how to effectively visualize better test-time scaling.

\bibliography{colm2025_conference}
\bibliographystyle{colm2025_conference}
\clearpage
\appendix
\section{Appendix}

\subsection{General Experimental Details}
\label{app:dataset_composition}
\paragraph{Pretraining dataset composition}

Our pretraining corpus consists of tokens drawn from diverse domains to ensure broad coverage of knowledge and reasoning capabilities. The majority of the data comes from the DCLM~\citep{li2025datacomplmsearchgenerationtraining} like baseline dataset and GitHub repositories~\citep{githubcode}. In addition, we include a range of specialized sources spanning mathematics, coding, scientific literature, and high-quality web content. Specifically, our mixture includes DeepMind Mathematics problems~\citep{saxton2019analysingmathematicalreasoningabilities}, Proof Pile 2 collections (ArXiv, Open Web Math, Algebraic Stack) from ~\citet{azerbayev2023llemma}, Stack Exchange from the pile~\citep{pile}, FineWeb-Edu~\citep{lozhkov2024finewebedu}, and smaller curated sets such as Natural Reasoning Dataset~\citep{yuan2025naturalreasoningreasoningwild28m} and AQuA~\citep{ling2017programinductionrationalegeneration}.

\paragraph{Pretraining Hyperparameters}
For temperature(T) in distillated pretraining, we do a grid-search over $T\in\{0.5,1,2,3\}$. We select the temperature which gives the best performance on standard benchmarks. In our experiments, $T=1$ worked the best. We pretrain with cosine scheduler using a learning rate of $3e^{-3}$ for 1B models and $3e^{-4}$ for 8B models.

\subsection{Proposition 1 (formal).} 
\setcounter{proposition}{0}
\begin{proposition} \hfill
\begin{itemize}[left=1pt]
\item If the number of sequences observed grow as $\mathcal{O}(k\log k + k \log (\frac{1}{\delta }))$, then the $\pi^{\mathsf{distill}} = \pi^{\mathsf{teacher}}$ with probability at least $1-\delta$. 
\item If the number of sequences observed grow as $\mathcal{O}\Big(\frac{(k\log k + (p/\epsilon^2-1)k  \log\log k)}{\delta}\Big)$, then for each row $i\in[k]$, $\mathbb{E}[\|\pi^{\mathsf{scratch}}_{i}-\pi_{i}\|_1] \leq \epsilon$ with probability $1-\delta$, where $\mathbb{E}$ is computed over the entire draw of the dataset.
\end{itemize}
\end{proposition}

\begin{proof}
To prove the first part, let us recollect a standard result.  

The coupon collector problem studies the following question.
Suppose each box contains a coupon, and there $k$ different types of coupons. What is the number of boxes we need to see $T$ before we have collected all $k$ coupons? Assuming each coupon is drawn uniformly at random, 

$$P(T> \beta k \log k ) < k^{-\beta +1 }$$

Substitute $\beta = 1 + \frac{\log \frac{1}{\delta}}{\log k}$, we obtain 

$$P(T>  k \log k  + k \log(\frac{1}{\delta})) < \delta$$

Translated to our setting, this means if we observe  $k \log k  + k \log(\frac{1}{\delta})$, then with probability at least $1-\delta$ each of the distinct $k$ tokens have been observed at the first position in the sequence. This completes the proof for the first part.

We now turn to the model trained from scratch. The log-likelihood of a model is written as $\sum_{ij} n_{ij}\log(\hat{\pi}_{ij})$, where $n_{j}$ is the number of times we see a token $j$ appear after token $i$. The solution to maximum likelihood is simply $\hat{\pi}_{ij} = \frac{n_{ij}}{n_i}$, where $n_i = \sum_{j\in [k]} n_{ij}$. $\hat{\pi}_{ij}$ is an unbiased estimator of $\pi_{ij}$. Define

For this model, we need to ensure that each row in the estimated matrix is close to the true row. Next, we want to bound the distance between $\|\hat{\pi}_{i,:}- \pi_{i,:}\|_1$, where we particularly use $\ell_1$
distance to emphasize the role of sparsity. Observe that the variance of each element of the row is $\mathbb{E}[(\hat{\pi}_{ij} - \pi_{ij})^2] = \frac{\pi_{ij}(1-\pi_{ij})}{\sum_{j}n_{ij}}$.

Observe that 
\begin{equation}
   \Big(\mathbb{E}[|\hat{\pi}_{ij}- \pi_{ij}|]\Big)^2 \leq  \mathbb{E}[(\hat{\pi}_{ij}- \pi_{ij})^2] = \frac{\pi_{ij}(1-\pi_{ij})}{\sum n_{ij}} \implies \mathbb{E}[|\hat{\pi}_{ij}- \pi_{ij}|] \leq \sqrt{ \frac{\pi_{ij}(1-\pi_{ij})}{\sum n_{ij}}} 
\end{equation}

To compute, $\|\hat{\pi}_{i}- \pi_{i}\|_1$, we only need to sum over the terms that are non-zero owing to the sparsity assumption. Suppose that without loss of generality first $p$ terms are non-zero. Hence, we obtain 

\begin{equation}
    \mathbb{E}[\|\hat{\pi}_{i}- \pi_{i}\|_1] = \sum_{j \leq p} \mathbb{E}[|\hat{\pi}_{ij}- \pi_{ij}|] \leq \sum_{j\leq p }\sqrt{ \frac{\pi_{ij}(1-\pi_{ij})}{\sum n_{ij}}}
    \label{eqn: l1_distance_bound}
\end{equation}

We can arrive at a simple upper bound for $\sum_{j \leq p}\sqrt{\pi_{ij}(1-\pi_{ij})}$ as follows. We again apply Cauchy-Schwarz inequality. We express  $$\sum_{j \leq p}\sqrt{\pi_{ij}(1-\pi_{ij})} = \braket{1, [\sqrt{\pi_{i1}(1-\pi_{i1})}, \sqrt{\pi_{i2}(1-\pi_{i2})}, \cdots, \sqrt{\pi_{ip}(1-\pi_{ip})}]}$$

$$\braket{1, [\sqrt{\pi_{i1}(1-\pi_{i1})}, \sqrt{\pi_{i2}(1-\pi_{i2})}, \cdots, \sqrt{\pi_{ip}(1-\pi_{ip})}]} \leq \sqrt{p}\sqrt{\sum_{j} (\pi_{ij})(1-\pi_{ij})} \leq \sqrt{p}$$

We substitute this in \eqref{eqn: l1_distance_bound} to obtain

\begin{equation}
    \mathbb{E}[\|\hat{\pi}_{i}- \pi_{i}\|_1] \leq \sum_{j\leq p }\sqrt{ \frac{\pi_{ij}(1-\pi_{ij})}{n_i}}  \leq \sqrt{\frac{p}{n_i}} 
\end{equation}

From the above, we can observe that if $n_i = \frac{p}{\epsilon^2}$, then 
$$\mathbb{E}[\|\hat{\pi}_{i}- \pi_{i}\|_1] \leq  \epsilon, \forall i \in [k]$$

Hence, if each token $i$ is observed at the first position of the sequence at least $\frac{p}{\epsilon^2}$, then we should obtain the desired outcome we set out to prove in this part. 

We now recollect the generalized version of coupon collector's problem. In the generalized version one is interested in computing the number of boxes to collect defined as $T_m$ before collecting $m$ copies of each coupon. In this case, 

$$\mathbb{E}[T_m] \approx k \log k + (m-1)k \log \log k$$

If we apply Markov inequality on the above, we obtain a simple bound 

$$P\Big(T_{m} \geq  \frac{1}{\delta} \cdot \mathbb{E}[T_m] \Big) \leq \delta $$

Thus from the above, we gather that if the number of boxes collected is at least $\frac{1}{\delta} \cdot \Big( k \log k + (m-1)k \log \log k\Big)$, then with probability at least $1-\delta$ we have collected $m$ copies of each coupon. 

We can now substitute $m=\frac{p}{\epsilon^2}$ to obtain our bound of $\frac{k\log k + (p/\epsilon^2-1)k  \log\log k)}{\delta}$. This completes the proof. 
\end{proof}

\subsection{Experimental Details for Bigram Sandbox and Induction Head Learning}
\label{app:bigram_experiment_details}

Our bigram sandbox experiments were designed to provide a simple, controlled testbed for understanding how distillation influences test-time scaling and in-context learning. All results in Section~\ref{sec:bigram} are derived from this setup.

\paragraph{Data generation.} The vocabulary consists of $k = 64$ tokens. The bigram transition matrix $\pi \in \mathbb{R}^{k \times k}$ was constructed to include a mix of low-, medium-, and high-entropy rows: low-entropy rows concentrated probability mass on 3–5 tokens; high-entropy rows were nearly uniform; medium-entropy rows had an intermediate profile. Trigger tokens were randomly selected (5, 10, or 20 triggers per experiment), with trigger-output mappings varying across sequences to induce induction head learning (following \cite{bietti2023birth}). Sequences were generated using a first-order Markov chain with these bigram transitions, with special logic to ensure copying behavior for trigger tokens.

\paragraph{Models.} Both teacher and student models were implemented as small Transformers with 2–4 layers, causal masking, and a fixed sequence length of 64. Teacher models used 128-dimensional embeddings; students used 64-dimensional embeddings. Training was performed with Adam optimizer and a cosine learning rate schedule.

\paragraph{Training.} Teacher models were trained on datasets of size 16k sequences. Student models were trained with either cross-entropy (CE) loss or knowledge distillation (KD), using soft logits from the teacher. Dataset sizes for students were 8k sequences i.e. half the data. The KD objective used temperature $T = 2.0$ and mixing coefficient $\alpha = 0.5$(Equation~\ref{eqn:erm_distillation}).

\paragraph{Evaluation.} All models were evaluated on a fixed held-out dataset of 4k sequences. Metrics included: Induction head accuracy (trigger $\rightarrow$ copy) as shown in Figure~\ref{fig:title_base_model_passk}c; and KL-divergence between the ground-truth distribution (bigram rows) and the learnt distribution for low-, medium-, and high-entropy rows as shown in Figure~\ref{fig:bigram_and_bayes_plots}.

Our full codebase will be released for reproducibility.

\subsection{Token Routing for mitigating drop in ICL with distillation}
\label{app:token_routing}
\begin{figure}[t!]
    \centering
    \begin{subfigure}{0.235\textwidth}
        \centering
        \includegraphics[width=\linewidth]{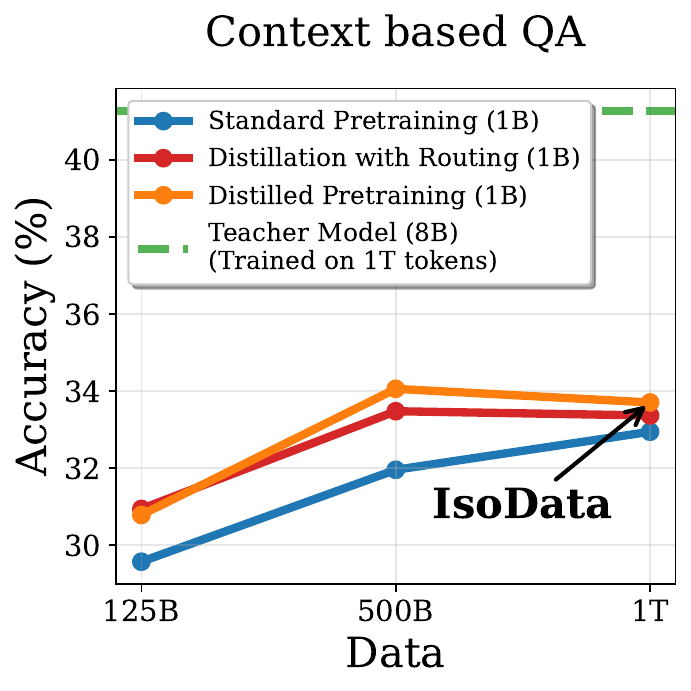}
    \end{subfigure}
    \hfill
    \begin{subfigure}{0.245\textwidth}
        \centering
        \includegraphics[width=\linewidth]{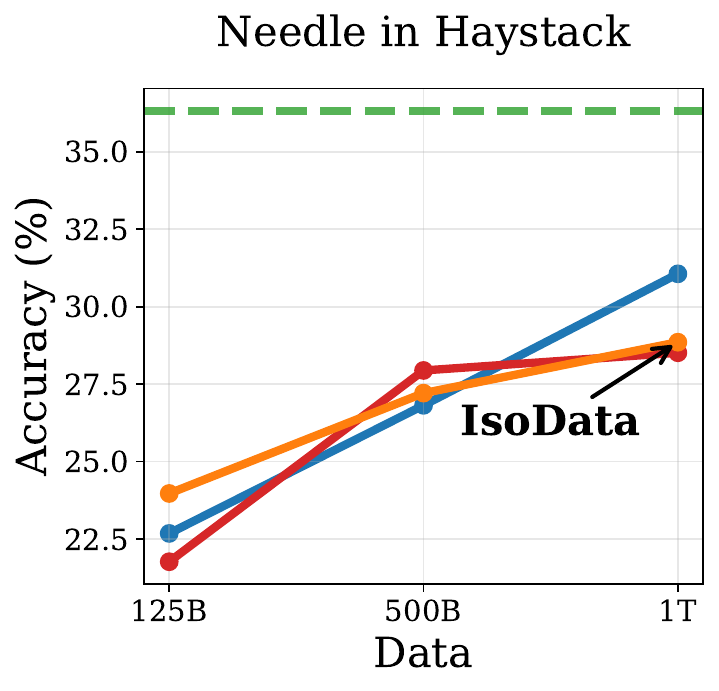}
    \end{subfigure}
    \hfill
    \hfill
    \begin{subfigure}{0.235\textwidth}
        \centering
        \includegraphics[width=\linewidth]{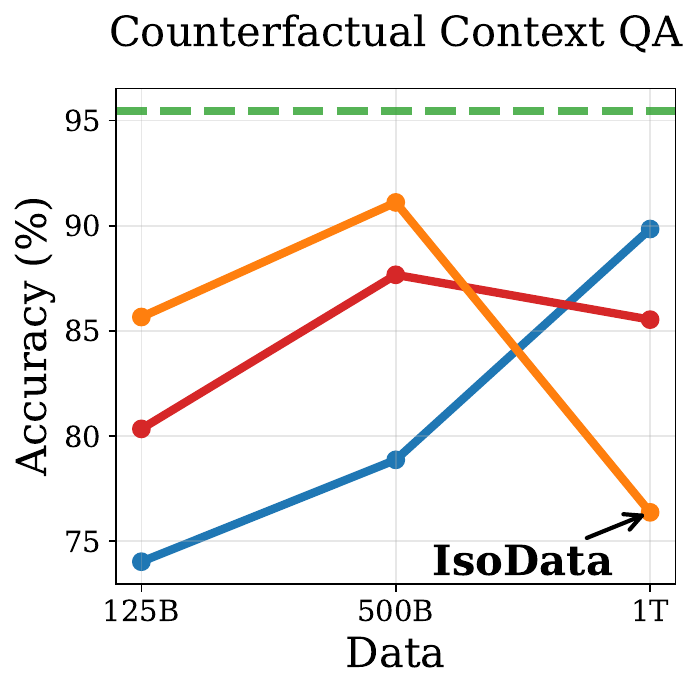}
    \end{subfigure}
    \hfill
    \begin{subfigure}{0.245\textwidth}
        \centering
        \includegraphics[width=\linewidth]{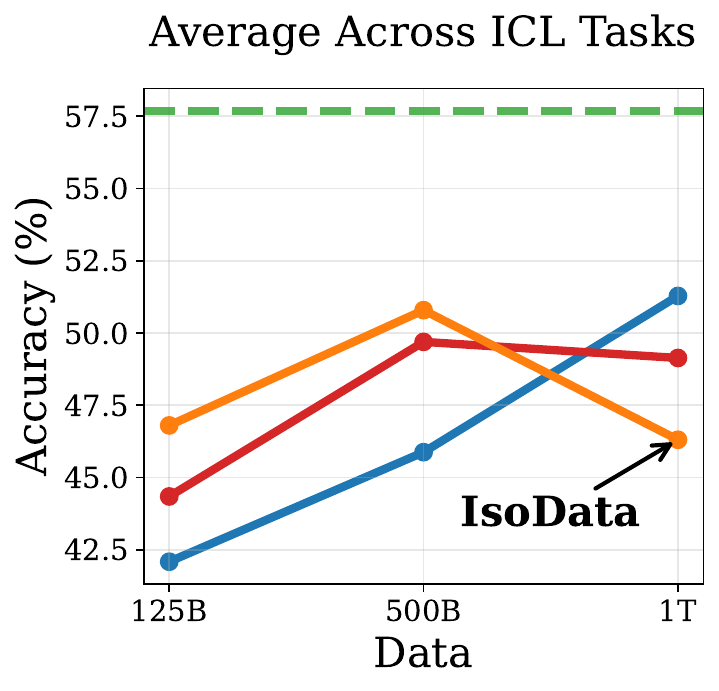}
    \end{subfigure}
    \caption{\textbf{Token Routing: Mitigating the Drop in In-Context Learning} In Figure~\ref{fig:entropy_routing_15p} we presented results when routing 15\% of the tokens. Here we present results when routing 30\% of tokens. 
    }
    \label{fig:entropy_routing_30p}
\end{figure}

We introduced token routing in \S~\ref{sec:token_routing} as a simple yet effective strategy to mitigate the drop in in-context learning observed with distilled pretraining. In Figure~\ref{fig:entropy_routing_15p}, we showed results when distillation loss is skipped on $x=15\%$ of the tokens in each sequence—specifically, those with the lowest entropy in the teacher’s soft labels. This routing improves in-context learning on 2 out of the 3 evaluated benchmarks. 
In Figure~\ref{fig:entropy_routing_30p}, we first share additional results when routing 30\% of the tokens. We observe that 30\% token routing improves performance only on 1 task compared to the 2 tasks when routing 15\% tokens. Moreover, too much token routing can hurt performance on standard tasks as shown in Table~\ref{tab:entropy_routing_additional_evals}.

We share the performance with token routing on standard language modeling tasks and reasoning benchmarks in Table~\ref{tab:entropy_routing_additional_evals}. We observe that routing 15\% of the tokens preserves the performance on standard language modeling benchmarks. However, if we further increase the tokens on which distillation is not performed to 30\%, there is a drop in performance, although it still remains above standard pretraining as one would expect.

{\large
\begin{table}
\setlength{\tabcolsep}{5pt} %
\renewcommand{\arraystretch}{2.5} %
\resizebox{\textwidth}{!}{%
\begin{tabular}{@{}cccccccccccccccc@{}}
\toprule
Run                           & \begin{tabular}[c]{@{}c@{}}HellaSwag\\ (Accuracy)\end{tabular} & \begin{tabular}[c]{@{}c@{}}TQA\\ (F1)\end{tabular} & \begin{tabular}[c]{@{}c@{}}MBPP\\ (\passone{})\end{tabular} & \begin{tabular}[c]{@{}c@{}}MBPP\\ (\passsixteen{})\end{tabular} & \begin{tabular}[c]{@{}c@{}}HumanEval+\\ (\passone{})\end{tabular} & \begin{tabular}[c]{@{}c@{}}GPQA\\ (EM)\end{tabular} & \begin{tabular}[c]{@{}c@{}}GSM8K\\ (\passone{})\end{tabular} & \begin{tabular}[c]{@{}c@{}}GSM8K\\ (\passsixteen{})\end{tabular} & \begin{tabular}[c]{@{}c@{}}ARC-C\\ (Acc.)\end{tabular} & \begin{tabular}[c]{@{}c@{}}ARC-E\\ (Acc.)\end{tabular} & \begin{tabular}[c]{@{}c@{}}COPA\\ (Acc.)\end{tabular} & \begin{tabular}[c]{@{}c@{}}MATH\\ (\passone{})\end{tabular} & \begin{tabular}[c]{@{}c@{}}MATH\\ (\passsixteen{})\end{tabular} & \begin{tabular}[c]{@{}c@{}}SQuAD\\ (F1)\end{tabular} & Avg.  \\ \midrule
NTP                           & 64.67                                                          & 29.74                                              & 14.78                                                   & 43.9                                                     & 9.76                                                          & 13.39                                               & 4.32                                                     & 31.92                                                     & 37                                                     & 65.79                                                  & 76                                                    & 2.2                                                     & 14.4                                                     & 51.38                                                & 32.80 \\
Distillation                  & 65.64                                                          & 33.68                                              & 17.03                                                   & 47.64                                                    & 9.76                                                          & 9.15                                                & 4.25                                                     & 33.59                                                     & 38.88                                                  & 66.55                                                  & 79                                                    & 0.6                                                     & 15.6                                                     & 55.34                                                & 34.05 \\
Distillation + Routing (15\%) & 65.58                                                          & 32.23                                              & 17.18                                                   & 45.91                                                    & 9.15                                                          & 9.38                                                & 5                                                        & 32.9                                                      & 38.71                                                  & 67.57                                                  & 79                                                    & 1.2                                                     & 16.4                                                     & 55.48                                                & 33.98 \\
Distillation + Routing (30\%) & 66.43                                                          & 30.56                                              & 17.35                                                   & 47.23                                                    & 6.71                                                          & 12.05                                               & 4.47                                                     & 32.98                                                     & 40.34                                                  & 68.54                                                  & 77                                                    & 0.6                                                     & 12                                                       & 52.77                                                & 33.50 \\ \bottomrule
\end{tabular}
}
\caption{\textbf{Token Routing (\S~\ref{sec:token_routing}) does not significantly hurt performance on standard benchmarks.} 
Doing distillation only on tokens for which teacher label has a high-entropy mitigates the drop in ICL performance (Figure~\ref{fig:entropy_routing_15p}) while preserving the performance on standard language modeling tasks and reasoning tasks, as shown in the table. This also reinforces the fact that gains in reasoning tasks come primarily from tokens where teacher label has high-entropy, and removing the distillation loss term for tokens where teacher label has low-entropy does not hurt standard tasks. As expected, routing a lot of tokens (e.g., 30\%) hurts the standard benchmark performance.}
\label{tab:entropy_routing_additional_evals}
\end{table}
}

\subsection{Additional Evaluations}
\paragraph{Evaluations for the 1B base models trained using Llama-3.1-8B as teacher}
We share additional evaluations on standard benchmarks for the base models in Table~\ref{tab:additional_base_model_evals}.

\begin{figure}[t!]
    \centering
    \begin{subfigure}{0.32\textwidth}
        \centering
        \includegraphics[width=\linewidth]{figures/pass_at_k_mbpp.pdf}
        \caption{}
    \end{subfigure}
    \hfill
    \begin{subfigure}{0.32\textwidth}
        \centering
        \includegraphics[width=\linewidth]{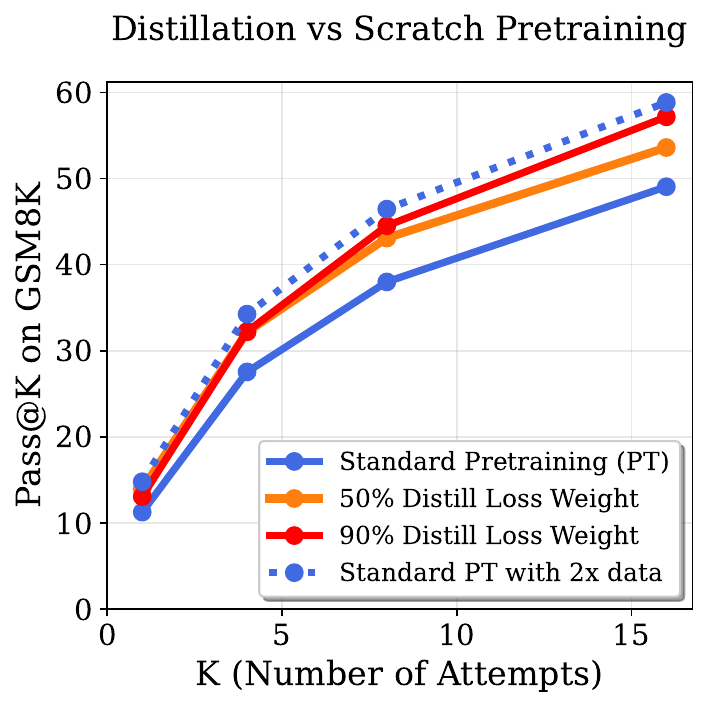}
        \caption{}
    \end{subfigure}
    \hfill
    \begin{subfigure}{0.32\textwidth}
        \centering
        \includegraphics[width=\linewidth]{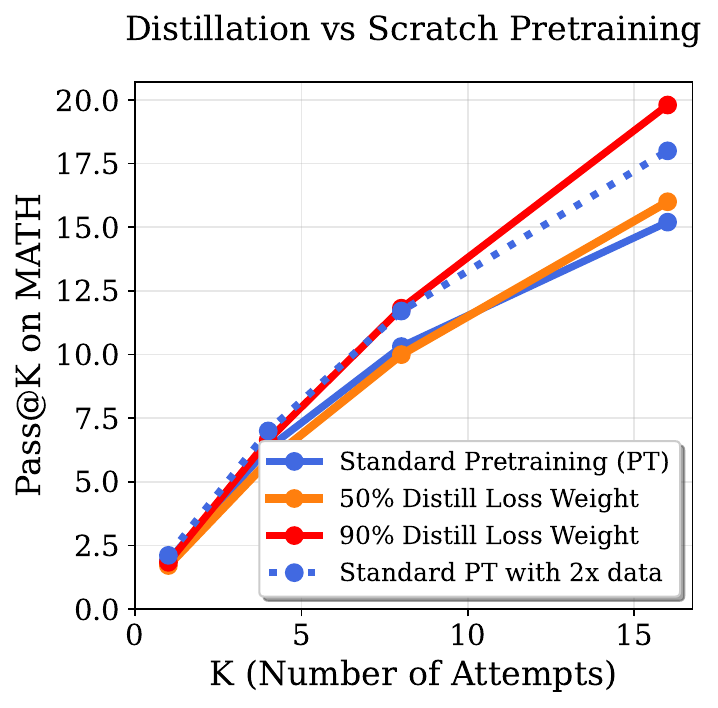}
        \caption{}
    \end{subfigure}
    \caption{\textbf{Distillation pretraining diversity leads to better post-training test-time scaling as well: }\textbf{(a)} Base model evaluations on coding task of most basic python problems (mbpp). Distillation pretrained models exhibit much stronger test-time scaling and diversity in generations, as exhibited by a higher \passk{} than even a model trained on $2\times$ more data with standard pretraining. Note that this is despite the fact that both models have a similar \passone. See Figure~\ref{fig:title_base_model_passk} for more tasks. \textbf{(b,c)} Diversity gains in base model evaluations persist even after post-training, as depicted by better test-time scaling after post-training as well on MATH and GSM8k.}
    \label{fig:post_train_pass_at_k}
\end{figure}

\paragraph{Base model diversity gains persist even after post-training}
Recall that in \S~\ref{sec:distillation_improves_diversity} and Figure~\ref{fig:test_time_scaling_main} we showed that distillation pretraining enables much more effective test-time scaling. In Figure~\ref{fig:post_train_pass_at_k} we show that these base model gains persist even after post-training these models on high quality reasoning data using off-policy distillation~\citep{yang2025qwen3technicalreport, guha2025openthoughtsdatarecipesreasoning,muennighoff2025s1simpletesttimescaling}.

\begin{figure}[t!]
  \centering
  \includegraphics[width=\textwidth]{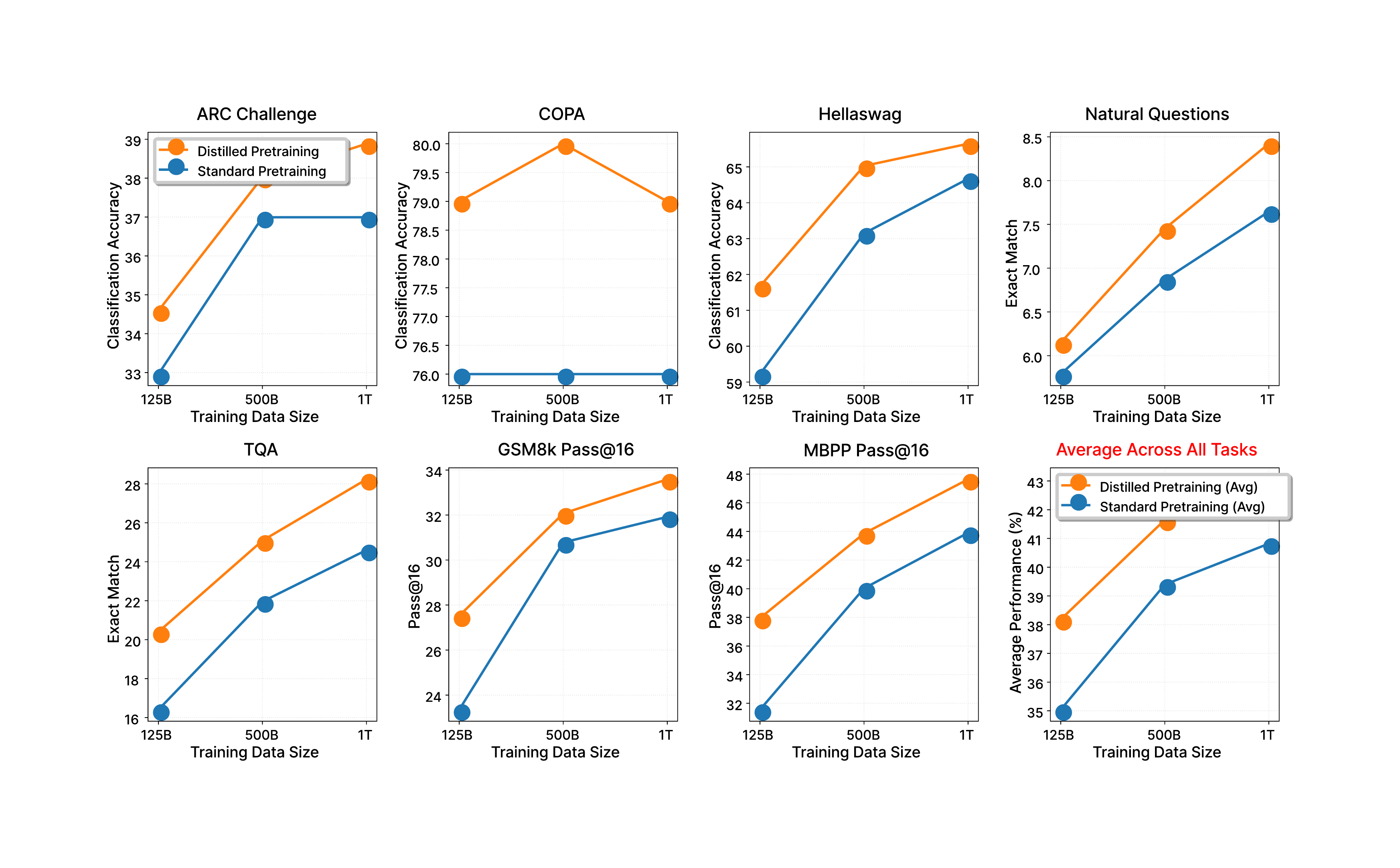}
\caption{\textbf{Distilled pretraining consistently outperforms standard pretraining even in IsoData setting()\S~\ref{sec:isodata}: } Unlike in-context learning and induction head tasks where distillation underperforms in the isodata regime (Figure~\ref{fig:isodata_icl}), distilled pretraining continues to yield better results on standard language modeling tasks that do not rely on induction heads—even when student models are trained on the full 1T tokens as used by the teacher. Moreover, we continue to see that distilled pretraining rewards with better test-time scaling on the GSM8k and MBPP plots (both as Pass@16 curves).}
  \label{fig:isodata_iwl_full}
\end{figure}

\paragraph{Additional evaluations for IsoData Models (trained using 8B param 1T token teacher)}
Recall that in \S~\ref{sec:icl} and Figure~\ref{fig:isodata_icl} we showed how distillation impairs in-context learning, especially in the ``IsoData'' setting where the teacher, student and the standard pretrained model all see the same data. Note that this is in \emph{stark contrast} with performance on standard language modeling tasks where the performance of distilled models continues to be better than standard pretrained models even under the isodata setting, as shown in Figure~\ref{fig:isodata_iwl_full}.

{\large
\begin{table}[]
\setlength{\tabcolsep}{5pt} %
\renewcommand{\arraystretch}{2.5} %
\resizebox{\textwidth}{!}{%
\begin{tabular}{@{}cccccccccccccccc@{}}
\toprule
Run                                                                 & \begin{tabular}[c]{@{}c@{}}HellaSwag \\ (Accuracy)\end{tabular} & \begin{tabular}[c]{@{}c@{}}TQA \\ (EM)\end{tabular} & \begin{tabular}[c]{@{}c@{}}MBPP \\ (\passone{})\end{tabular} & \begin{tabular}[c]{@{}c@{}}MBPP \\ (\passsixteen{})\end{tabular} & \begin{tabular}[c]{@{}c@{}}HumanEval+ \\ (\passone{})\end{tabular} & \begin{tabular}[c]{@{}c@{}}GPQA \\ (EM)\end{tabular} & \begin{tabular}[c]{@{}c@{}}GSM8K \\ (\passone{})\end{tabular} & \begin{tabular}[c]{@{}c@{}}GSM8K \\ (\passsixteen{})\end{tabular} & \begin{tabular}[c]{@{}c@{}}ARC-C \\ (Acc.)\end{tabular} & \begin{tabular}[c]{@{}c@{}}ARC-Easy \\ (Acc.)\end{tabular} & \begin{tabular}[c]{@{}c@{}}COPA \\ (Acc.)\end{tabular} & \begin{tabular}[c]{@{}c@{}}MATH \\ (\passone{})\end{tabular} & \begin{tabular}[c]{@{}c@{}}MATH \\ (\passsixteen{})\end{tabular} & \begin{tabular}[c]{@{}c@{}}SQuAD \\ (EM)\end{tabular} & Avg.  \\ \midrule
Standard  PT                                                        & 59.22                                                           & 21.29                                                  & 7.85                                                     & 31.56                                                     & 7.32                                                           & 11.38                                                & 2.96                              & 23.35                              & 32.96                                                   & 60.97                                                      & 76.00                                                  & 1.00                             & 10.80                             & 45.38                                                 & 28.00 \\
\begin{tabular}[c]{@{}c@{}}50\% Distill loss weight\end{tabular} & 61.13                                                           & 24.48                                                 & 12.78                                                    & 41.00                                                     & 5.49                                                           & 12.05                                                & 2.20                              & 26.00                              & 34.51                                                   & 62.37                                                      & 81.00                                                  & 0.80                             & 14.60                             & 45.21                                                 & 29.92 \\
\begin{tabular}[c]{@{}c@{}}90\% Distill loss weight\end{tabular}  & 62.07                                                           & 25.55                                                  & 13.56                                                    & 40.88                                                     & 8.54                                                           & 12.05                                                & 3.11                                                      & 26.99                                                      & 34.42                                                   & 62.28                                                      & 81.00                                                  & 1.60                                                     & 17.40                                                     & 53.35                                                 & 31.24 \\
\begin{tabular}[c]{@{}c@{}}Standard PT with 2x data\end{tabular} & 61.08                                                           & 23.79                                                 & 11.94                                                    & 39.56                                                     & 7.93                                                           & 8.04                                                 & 4.40                                                      & 26.97                                                      & 35.45                                                   & 63.00                                                      & 76.00                                                  & 2.40                                                     & 16.30                                                     & 52.22                                                 & 30.00 \\ \bottomrule
\end{tabular}
}
\caption{Additional evaluations for the 1B base models trained on 125B tokens ($1\times$ data) used in this paper. One can observe the better test-time scaling properties exhibited by distillation pretrained models, on MATH and GSM8k. \passone{} is lower compared to standard pretrained model, but \passsixteen{} is higher.}
\label{tab:additional_base_model_evals}
\end{table}
}

\end{document}

%% file: macros.tex
\newcommand{\passsixteen}{$\mathsf{pass}@16$}
\newcommand{\passone}{$\mathsf{pass}@1$}
\newcommand{\passk}{$\mathsf{pass}@k$}
\newcommand{\spt}{SPT }
\newcommand{\sptd}{SPT-2x }

\newcommand{\dptf}{DPT-50 }
\newcommand{\dptn}{DPT-90 }